\pdfoutput=1

\PassOptionsToPackage{table}{xcolor}
\documentclass[11pt]{article}

\usepackage[final]{acl}

\usepackage{times}
\usepackage{latexsym}

\usepackage[T1]{fontenc}

\usepackage[utf8]{inputenc}

\usepackage{microtype}

\usepackage{inconsolata}

\usepackage{graphicx}
\usepackage{balance}       
\usepackage{graphics}      
\usepackage{hyperref}
\usepackage{color}
\usepackage{booktabs}
\usepackage{textcomp}
\usepackage{subcaption}
\usepackage{enumerate}
\usepackage{xcolor}
\usepackage{lipsum}
\usepackage{makecell}
\usepackage{multicol}
\usepackage{multirow}
\usepackage{array}
\usepackage{verbatimbox}
\usepackage{enumitem}
\usepackage{amsmath}
\usepackage{stfloats}
\usepackage{graphicx}
\usepackage{amsthm}
\usepackage{listings}
\usepackage{caption} 
\usepackage[export]{adjustbox}
\usepackage{xspace}
\usepackage{epsfig}
\usepackage[linesnumbered]{algorithm2e}
\usepackage{algpseudocode}
\usepackage{tabularx}
\usepackage{arydshln}
\usepackage[bottom]{footmisc}
\usepackage{tcolorbox}
\usepackage{stackengine}
\usepackage{placeins}

\newcolumntype{L}[1]{>{\raggedright\let\newline\\\arraybackslash\hspace{0pt}}m{#1}}
\newcolumntype{C}[1]{>{\centering\let\newline\\\arraybackslash\hspace{0pt}}m{#1}}
\newcolumntype{R}[1]{>{\raggedleft\let\newline\\\arraybackslash\hspace{0pt}}m{#1}}

\makeatletter
\def\thickhline{%
  \noalign{\ifnum0=`}\fi\hrule \@height \thickarrayrulewidth \futurelet
   \reserved@a\@xthickhline}
\def\@xthickhline{\ifx\reserved@a\thickhline
               \vskip\doublerulesep
               \vskip-\thickarrayrulewidth
             \fi
      \ifnum0=`{\fi}}
\makeatother

\makeatletter
\def\thickhlinespace{%
  \addlinespace[1ex]
  \noalign{\ifnum0=`}\fi\hrule \@height \thickarrayrulewidth \futurelet
   \reserved@a\@xthickhline
   \addlinespace[1ex]
   }
\def\@xthickhlinespace{\ifx\reserved@a\thickhline
               \vskip\doublerulesep
               \vskip-\thickarrayrulewidth
             \fi
      \ifnum0=`{\fi}}
\makeatother

\newlength{\thickarrayrulewidth}
\newlength\Origarrayrulewidth

\algnewcommand{\IfThenElse}[3]{
  \State \algorithmicif\ #1\ \algorithmicthen\ #2\ \algorithmicelse\ #3}

\definecolor{downredcolor}{HTML}{e31a1c}
\definecolor{upgreencolor}{HTML}{33a02c}

\definecolor{DarkGreen}{HTML}{5DAC81}

\definecolor{E+F}{RGB}{	255, 99, 71}
\definecolor{B+F}{RGB}{255, 165, 0}
\definecolor{E+I}{RGB}{	173, 216, 230}
\definecolor{B+I}{RGB}{	30, 144, 255}
\definecolor{D}{RGB}{	60, 179, 113}

\newcommand{\grayrow}{\rowcolor[gray]{0.9}}
\definecolor{maroon}{cmyk}{0,0.87,0.68,0.32}

\usepackage{tikz}
\newcommand{\CircleEF}[1][0.7]{%
    \tikz[baseline=(char.base)]\node[shape=circle,inner sep=2pt,line width=1pt,draw=gray,fill=E+F,scale=#1] (char) {\phantom{T}};\hspace{-1pt}
}

\newcommand{\CircleBF}[1][0.7]{%
    \tikz[baseline=(char.base)]\node[shape=circle,inner sep=2pt,line width=1pt,draw=gray,fill=B+F,scale=#1] (char) {\phantom{T}};\hspace{-1pt}
}

\newcommand{\CircleEI}[1][0.7]{%
    \tikz[baseline=(char.base)]\node[shape=circle,inner sep=2pt,line width=1pt,draw=gray,fill=E+I,scale=#1] (char) {\phantom{T}};\hspace{-1pt}
}

\newcommand{\CircleBI}[1][0.7]{%
    \tikz[baseline=(char.base)]\node[shape=circle,inner sep=2pt,line width=1pt,draw=gray,fill=B+I,scale=#1] (char) {\phantom{T}};\hspace{-1pt}
}

\newcommand{\CircleD}[1][0.7]{%
    \tikz[baseline=(char.base)]\node[shape=circle,inner sep=2pt,line width=1pt,draw=gray,fill=D,scale=#1] (char) {\phantom{T}};\hspace{-1pt}
}

\definecolor{darkgreen}{rgb}{0.0, 0.5, 0.0}
\definecolor{usercolor}{RGB}{200, 230, 250} 
\definecolor{coachcolor}{RGB}{180, 250, 180} 

\newcommand{\xhdr}[1]{\vspace{1.0mm}\noindent{{\bf #1.}}}

\title{Substance over Style: Evaluating Proactive \\ Conversational Coaching Agents}

\author{
 \textbf{Vidya Srinivas\textsuperscript{1}\thanks{Work done during an internship at Google}},
 \textbf{Xuhai Xu\textsuperscript{2}},
 \textbf{Xin Liu\textsuperscript{1,2}},
 \textbf{Kumar Ayush\textsuperscript{2}},
\\
 \textbf{Isaac Galatzer-Levy\textsuperscript{2}},
 \textbf{Shwetak Patel\textsuperscript{1,2}},
 \textbf{Daniel McDuff\textsuperscript{1,2}},
 \textbf{Tim Althoff \textsuperscript{1,2}},
\\
 \textsuperscript{1}University of Washington,
 \textsuperscript{2}Google Research
\\
 \small{
    \textit{vysri@cs.washington.edu, \{orson, xliucs, dmcduff, althoff\}@google.com}
 }
}

\NewDocumentCommand{\standout}{m}{\texttt{#1}}

\begin{document}
\maketitle

\begin{abstract}
 While NLP research has made strides in conversational tasks, many approaches focus on single-turn responses with well-defined objectives or evaluation criteria. In contrast, coaching presents unique challenges with initially undefined goals that evolve through multi-turn interactions, subjective evaluation criteria, and mixed-initiative dialogue. In this work, we describe and implement five multi-turn coaching agents that exhibit distinct conversational styles, and evaluate them through a user study, collecting first-person feedback on 155 conversations. We find that users highly value core functionality, and that stylistic components in absence of core components are viewed negatively. By comparing user feedback with third-person evaluations from health experts and an LM, we reveal significant misalignment across evaluation approaches. Our findings provide insights into design and evaluation of conversational coaching agents and contribute toward improving human-centered NLP applications.
 \end{abstract}

\section{Introduction}
\label{sec:introduction}
\begin{figure}[ht]
    \centering
    \includegraphics[width=0.45\textwidth]{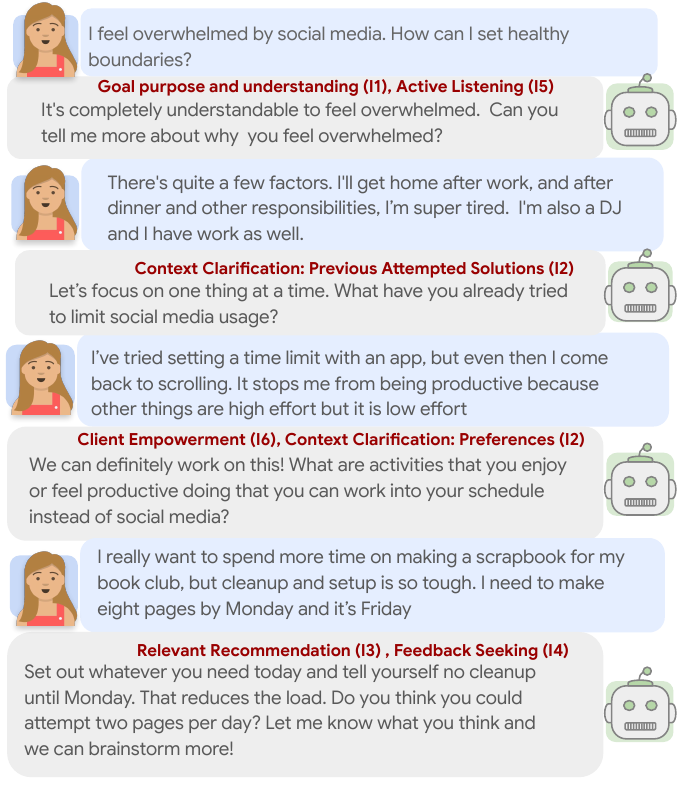} 
    \vspace{-4mm}
    \caption{\textbf{Example Coaching Conversation.} Different coaching behaviors are present throughout the conversation (highlighted in red) based on user responses, corresponding to recommended coaching behaviors (see \autoref{sec:he-interview}). Full transcripts in \autoref{app:conv-sample}}.
    \label{fig:introfigure}
    \vspace{-10mm}
\end{figure}

Recent NLP research has focused on evaluating tasks in conversational settings, including proactive information-seeking~\cite{pqa1}, multi-turn conversational dialogue~\cite{mt1, mt2, mt3}, and others. While significant progress has been made, the focus has largely been on exercises that represent one or multiple of the following properties: a single, unchanging, clearly communicated goal, a single correct answer, a high-quality answer expressed in a single turn or utterance, an assumption that objective quality can be evaluated by third parties (e.g., experts or language models), or a clear interaction structure (e.g., all user or model-initiated). However, proactive human-centered NLP applications such as personalized target-guided dialogues~\cite{iac} often exhibit several, concurrent, contrastive challenges. This work focuses on coaching conversations, which  differ from the aforementioned tasks with the presence of:
1) open-ended multi-turn interactions, 
2) initially ambiguous task definition that must be informed through multiple user interactions, 
3) multiple and changing goals that must be refined and prioritized,
4) potentially off-topic user input throughout long-form conversation,
5) varied conversational styles and preferences, 
6) mixed-initiative interactions, in which a coach must balance addressing user goals, needs and empowerment,
7) unspoken user needs and incomplete context, and the
8) absence of a single, objectively correct answer (instead relying primarily on subjective first-person evaluation).

As a result, coaching remains difficult and underexplored for proactive agents. In this work, we aim to 1) determine essential components of LM agent coaching, 2) describe and evaluate LM coaching agents that express distinct coaching styles (from common directive and interrogative to facilitative styles informed by human coaching experts), and 3) better understand agreement or discrepancies between first and third person evaluations of such conversations to inform future approaches. We use health coaching as an illustrative example to frame the coaching domain. Through our study, we find that questioning-only conversation styles typically reduce user engagement and satisfaction. We also find that users place high value on core functional components, which we refer to as substance, and perceive stylistic elements negatively when this substance is lacking. Finally, we find that LMs perform poorly as auto-raters for human-centered subjective tasks, but demonstrate reasonable agreement with users for objective human-centered tasks.
Our contributions are as follows:
\begin{itemize} [nosep, leftmargin=*]
    \item We synthesize the expertise and recommendations of health coaching experts ($N=11$) across a variety of coaching domains to define key capabilities for open-ended coaching.
    \item We identify LM coaching characteristics and create five conversational multi-turn coaching agents that incorporate different levels of these characteristics to embody directive, interrogative, and facilitative coaching styles
    \item We conduct a user study ($N=31$, 50 person hours of feedback, 155 conversations) where participants engage in unsupervised discussions on self-directed, open-ended coaching topics with these agents. We collect first-person feedback.
    \item We examine the alignment between first-person feedback and third-person annotations from health experts and an LM, highlighting shortcomings of evaluation approaches in this setting.
\end{itemize}


\section{Initial Health Expert Interviews}
\label{sec:he-interview}
Health coaches provide general guidance to a user who is looking to pursue a health goal. To inform the design of our coaching agents across general coaching domains, we conducted a formative study with 11 coaches across domains of sleep, fitness, and life coaching. These coaches, or experts, possessed advanced degrees in their respective fields, coupled with substantial professional experience ranging from 4 to 46 years. The one-hour interview consisted of two parts--–open-ended questions through which experts shared their expertise on coaching best-practices, and a case study, in which experts reviewed, turn-by-turn, a conversation between a real end-user and a single-prompt LM.

\subsection{Health Expert Insights}
\label{sub:he-insights}
Based on our interviews, we identified \textbf{conversational expert insights} which we classified into two categories: \textit{style} and \textit{substance}. \textit{Style} refers to how something is presented—-its form or delivery. This includes language, tone, and overall impression. \textit{Substance} refers to core components of a helpful and actionable coaching experience, including exploration, context gathering, goal-seeking, and relevant recommendation.
\begin{itemize}[nosep, leftmargin=*]
    \item \standout{I1 }\textbf{Goal and purpose understanding} (\textit{substance}): Understanding goals and motivations is important. Coaches maintain a goal-oriented conversation and bring clients back to the main goal when they deviate off-topic.
    \item \standout{I2 } \textbf{Context clarification} (\textit{substance}): Coaches collect and understand clients’ constraints, preferences, and prior resolution attempts to better personalize recommendations.
    \item \standout{I3 } \textbf{Relevant recommendations} (\textit{substance}): Coaches provide recommendations that are relevant, actionable, and context-sensitive. 
    \item \standout{I4 } \textbf{Feedback seeking} (\textit{substance}): Coaches solicit user feedback and update recommendations as necessary based on this feedback.
    \item \standout{I5 }\textbf{Active listening} (\textit{style}): Coaches provide occasional reflections to ensure correct understanding and goal alignment promoting client engagement and trust.
    \item \standout{I6 } \textbf{Client empowerment} (\textit{style}): Coaches build trust with their clients and guide them to navigate their own paths when exploring potential solutions. Coaches are supportive when clients are moving in a good direction.
\end{itemize}

\subsection{Turn-by-Turn Expert Evaluation}
The single prompt used to generate example conversations for the turn-by-turn part of the health expert interviews contained both control flow decisions (i.e. when to change topic, when to ask questions, etc.) and coaching instructions. From expert annotations, we identified cases of premature recommendation and fixated questioning, in which the LM either made recommendations before seeking relevant information or continued down a single, narrow line of questioning without considering additional factors (expert quotes available in \autoref{app:turnbyturn-quotes}). These health expert interviews revealed consistent, significant shortcomings of static LM prompting approaches. These issues, along with the expert insights, informed our designs of conversational flow (\autoref{sub:exp-control}).

\section{Conversational Paradigms}
\label{sec:conv-paradigms}
Before developing a coaching agent that embodies health expert insights, we recognize two conversational paradigms that already exist between humans and LMs, and point out how \textit{expert insights emphasize a third, distinct form of conversation}. 

\textit{Interrogative} interaction occurs when either a human or an LM provides one-sided questioning in a conversation while the opposite party only provides answers; a recommendation is potentially made at the end of the conversation by the questioning party. Such an approach maximizes knowledge gain but minimizes interaction. Questionnaires have always been used for general information seeking, but with the introduction of LMs to interpret and understand responses,  proactive QA for open-ended information seeking often embodies this paradigm~\cite{interrogative1, interrogative3, interrogative2}. 

\textit{Directive} interaction is a paradigm in which the LM takes an active role in providing constant solutions, instructions, and feedback to help a user achieve a goal. The most common examples are today's instruction-tuned LMs, such as ChatGPT, Claude, and Gemini~\cite{gemini}, among others. These models strive to provide continuous guidance and recommendations to a user. This is also a paradigm in coaching~\cite{directive1, directive2}.

\textit{Facilitative} interaction is a paradigm in which an LM focuses on guiding individuals towards their own solutions rather than providing immediate answers or recommendations. This interaction mode is emphasized in coaching~\cite{facilitative1, faciltative2}, but is not inherent to today's LMs.  Through our formative study \autoref{sec:he-interview}, we identify a central tension in open-ended conversational scenarios, specifically between making a recommendation and gathering further context from the user~\cite{qrec-tension}.

\section{Design Motivation and Components}
\label{sec:dm-comp}
We introduce intended LM characteristics that can be varied to create different coaches. These are not to be confused with \textit{expressed} or \textit{perceived} characteristics. \textit{Expressed} characteristics are those that are evaluated by a third party. We show alignment of expressed and intended characteristics in \autoref{sub:coach-expected}. \textit{Perceived} characteristics, are evaluated, post-conversation, in \autoref{sec:evaluation-user}. This differentiation can be critical for conversational constructs, including empathy~\cite{empathy1, empathy2}. Two modules encode different levels of coaching expertise and three agentic conversational flow modules allow us to implement the paradigms in \autoref{sec:conv-paradigms}. All agents use Gemini 1.5 Pro as a base LM.

\subsection{Coaching Expertise Variations}
\label{sub:coaching-variations}
We introduce two coaching modules, \standout{Base}, and \standout{Expert}, which vary \textit{style} components as defined in \autoref{sub:he-insights}. The \standout{Base} and \standout{Expert} modules share core instructions for proactive questioning; however, they differ in specific conversation objectives and conversation flow instructions. One can think of the \standout{Base} module as a coach with good intention but without formal training. In particular, the \standout{Base} module specifically outlines proactive questioning guidelines, but does not define specific topics, such as constraints or preferences, to talk about or heavily emphasize motivational language (see \autoref{app:prompts} for all prompts).

In contrast, one can think of the \standout{Expert} module as an experienced coach. The \standout{Expert} module explicitly outlines a goal-oriented questioning line, first breaching goal and purpose, and then constraints, preferences, and barriers before making a recommendation. The \standout{Expert} module  (see \autoref{app:prompts} for prompt) emphasizes motivational behavior, explicitly incorporating active listening (\standout{I5}) and client empowerment (\standout{I6}).

 \subsection{Conversation Flow Variations}
\label{sub:exp-control}
We introduce \textit{interrogative}, \textit{directive}, and \textit{facilitative} conversation flows, which vary in \textit{substance} components as defined in \autoref{sub:he-insights}. We first observe that the \textit{directive} flow is trivial to implement through a simple, single prompt. Our base LM, Gemini, inherently embodies a recommendation-heavy, suggestion-first flow (\standout{I3)} as it is instruction-tuned to do so. It does not place any explicit emphasis on exploratory behavior (\standout{I1}), context gathering (\standout{I2)}, or feedback seeking (\standout{I4}). 

While some coaching styles are easily elicited through simple prompts, others are harder to implement because it is difficult to specify through a single prompt when to make control flow decisions. These decisions include factors such as when to stop questioning, when to make a recommendation, or when to end a conversation. Deciding when to make such decisions often requires further, parallel control decisions to be considered. To make these decisions, we introduce explicit conversation flow, which defines a chain of thought~\cite{cot} before each agent response.

Explicit conversation flow modules steer conversation direction, executing control decisions in parallel during each turn. Each module consists of two LM inferences using Gemini 1.5 Pro. The first inference runs after each user utterance and outputs a binary decision. The second inference specifies how to perform the LM response to the user and executes based on this binary decision. This allows all first inferences to be run in parallel, while only one second inference will execute. When multiple first inferences return positive decisions, the first module in the conversation flow is prioritized (questioning before recommendation etc.). This allows the furthering of conversation to be prioritized over conclusion of the conversation.

The \standout{Probing} module is introduced to ask further clarification questions when something is unclear, vague, or when the user seems unsure or confused. The \standout{Recommendation} module is introduced to determine when to make a recommendation and to seek user feedback. The \standout{Resolution} module is introduced to determine when a conversation has reached a reasonable conclusion. All prompts are presented in \autoref{app:prompts}. We use a combination of the \standout{Probing} and \standout{Recommendation} modules to implement the \standout{Interrogative} conversational flow (emphasizing context gathering (\standout{I2}), and de-emphasizing goal and purpose understanding (\standout{I1}), feedback seeking (\standout{I4}), and suggestion frequency (\standout{I3})) and a combination of the \standout{Recommendation} and \standout{Resolution} agents to implement the \standout{Facilitative} conversational flow, emphasizing goal and purpose understanding (\standout{I1}), context gathering (\standout{I2}), and feedback seeking (\standout{I4}), and exhibiting medium suggestion frequency (\standout{I3})). We discuss how these styles can be combined to produce LM coaches in \autoref{sec:agent-design}.

\section{Coaching Agent Design}
\label{sec:agent-design}
Using these components we design five agents. The \standout{Base-Interrogative} and \standout{Expert-Interrogative} agents are implemented with one coaching module, \standout{Base} or \standout{Expert} (\autoref{sub:coaching-variations}), and \standout{Interrogative} conversational flow. The coaching module controls foundational proactive coaching while the \standout{Interrogative} conversational flow emphasizes knowledge gain before a recommendation is made.

The \standout{Base-Facilitative} and \standout{Expert-Facilitative} agents are implemented with one coaching module, \standout{Base} or \standout{Expert}, and \standout{Facilitative} conversational flow. The coaching module controls foundational proactive coaching, while the \standout{Facilitative} flow enforces timely recommendation and  graceful conclusion. We implement the directive coaching style in the \standout{Directive} agent with a prompt that asks the LM to act as a conversational health coach. We do not additionally prompt this agent or add control flow as it already embodies the coaching paradigm that we mean to test in \autoref{sub:coaching-variations} and \autoref{sec:conv-paradigms}. 

\subsection{Hypotheses for Agent Behavior}
\label{sub:agent-behavior}
We make the following hypotheses about agent behavior based on intended agent behaviors and level of alignment with expert insights:
\begin{itemize}[nosep, leftmargin=*]
    \item \standout{H1} (\textbf{Style}):  
    Agents with an \textit{expert} coaching module will perform better than corresponding agents with a \textit{base} coaching module for active listening (\standout{I5}) and client empowerment (\standout{I6}), with the \standout{Directive} agent performing in between.

    \item \standout{H2} (\textbf{Substance}):  
    \standout{Facilitative} agents will perform better than \standout{Directive} agents, which will perform better than \standout{Interrogative} agents on the following: goal purpose and understanding (\standout{I1}), context clarification (\standout{I2}), relevant recommendations (\standout{I3}), and feedback seeking (\standout{I4}).

    \item \standout{H3} (\textbf{Overall}):  
    Overall, in terms of alignment with expert insights across all indicators (\standout{I1}--\standout{I6}), agent performance is expected to decrease in the following order (from best to worst): \standout{Expert-Facilitative}, \standout{Base-Facilitative}, \standout{Directive}, \standout{Expert-Interrogative}, \standout{Base-Interrogative}.

\end{itemize}

In \autoref{sec:user-study-res-disc}, we systematically evaluate these agents. In \autoref{sub:coach-expected}, we validate these hypotheses through expert evaluation.

\section{Research Questions \& Study Design}
\label{sec:evaluation-user}
In this section, we describe a user study evaluation of our five proposed coaching agents, to answer the following research questions:
\label{sub:rq}

\noindent \textbf{RQ1:} Which agents do users prefer overall and do they have specific characteristics that stand out?
\noindent \textbf{RQ2:} What matters more in a coaching conversational agent, \textit{substance} or \textit{style}?\\
\noindent \textbf{RQ3:} Are LMs competent coaches?\\
\noindent \textbf{RQ4:} What are the factors that drive user satisfaction across agents?\\
\noindent \textbf{RQ5:} Can LM auto-raters capture first-person user preferences and third-person coach expertise?
\noindent \textbf{RQ6:} Do agent LM coaches demonstrate intended behavior?

\subsection{End-User Evaluation Study} 
\label{sub:user-eval}
We recruited 37 English-speaking participants, 31 of whom participated in our study. Participants varied in age (18 in range 25-34, 12 in range 35-44, and 1 in range 55-64) and self-identified gender (19.3 \% female, 81.7\% male).
Each participant attended a 1.5-hour session, during which they interacted with and evaluated the five agents in an order denoted by a balanced Latin square. To guide interactions, we curated 33 open-ended scenarios reflecting common concerns across sleep, fitness, repeated habits, and everyday situations (\autoref{tab:scenario-list}). Participants selected the scenario that best aligned with personal interests to ensure a realistic coaching use case. Participants were instructed to treat each agent interaction independently, providing consistent background information as if consulting a new expert. Each interaction began with the same initial sentence from the chosen scenario, after which participants conversed freely. Conversations concluded naturally—-either with a resolution or recommendation, or when further discussion was deemed unproductive. After each interaction, participants assessed the agent individually.  At the end of the study, the users ranked all five agents on different metrics, discussed further in \autoref{sec:results}.

A meaningful evaluation requires users to be invested in such a study beyond participation compensation. A key strength of our study is the flexibility it offers participants in scenario selection, ensuring relevance to personal behavioral health goals. Additionally, by allowing open-ended interactions without rigid guidelines on conversation structure, our study captures a range of realistic user experiences and human-language model interactions that may be overlooked in constrained studies or automated evaluations.

\section{User Study Results \& Discussion}
\label{sec:user-study-res-disc}

\label{sec:results}
In this section, we describe the results of our user study. We primarily collected three data types: Likert scale ratings, agent rankings, and open-ended feedback. All Likert scale ratings are on a scale of 1-5 (1--worst, 5--best). When applicable, quotes are specified along with participant number.

\begin{figure*}[t]
    \centering
    \vspace{-6mm}
    \includegraphics[width=\textwidth]{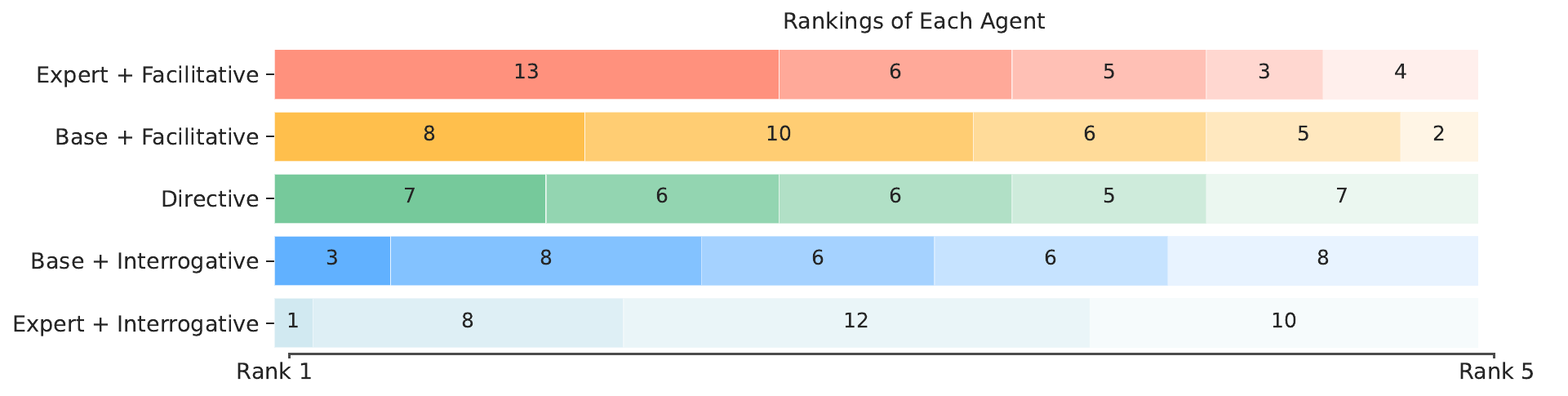} 
    \vspace{-8mm}
    \caption{\textbf{Overall Agent Rankings.} Users ranked all agents from best (Rank 1) to worst (Rank 5).}
    \label{fig:ranking-boxplot}
    \vspace{-2mm}
\end{figure*}

\subsection{Which agents do users prefer overall?} 
\label{sub:overall-pref}
\xhdr{Method} 
We define win rate as the combined percentages of first and second rankings of agents. We also collected participant ratings on a 5-point scale for self-identified LM chatbot usage experience and experience working with human health coaches. A rating of 1 indicated no experience; a rating of 5 indicated extensive experience.

\xhdr{Results} We observe that 67.7\% of users rank facilitative agents as their top preference, followed by directive, and then interrogative agents (\autoref{fig:ranking-boxplot}). The \standout{Expert-Facilitative} accounts for 41.9\% of first rankings and \standout{Base-Facilitative} accounts for 25.8\%. Win rates are significantly higher for \standout{Expert-Facilitative} and \standout{Base-Facilitative} agents, at 61.29\% and 58.06\% over the \standout{Directive} agent at 41.94\% and over \standout{Expert-Interrogative} and \standout{Base-Interrogative} agents at 35.48\% and 3.22\%.
Interrogative agents experienced more forced interaction endings than natural ones as compared to other agents: 58.1\% and 71.0\% of the time for \standout{Base-Interrogative} and \standout{Expert-Interrogative} as compared to the 16.1-19.4\%  for facilitative agents and 25.8\% for the directive agent.
Of users with moderate to extensive chatbot experience (3-5), 29.2\% preferred the \standout{Directive} agent as their top choice, while 0\% of users with lower levels of chatbot experience (1-2) preferred this agent as their top choice. We do not observe significant trends in agent preference based on prior experience with human health coaches.

\xhdr{Discussion} It may be useful to collect as much information about a user as possible before making a recommendation. Though interrogative agents do in fact sometimes generate more desirable conversation flows than the facilitative agents (\textit{``This agent ...seemed to be more action-oriented... The conversation more quickly went into solutions, which I preferred."} -P29), too much probing can give the impression that the agent has lost sight of the original goals (\textit{``It deviated from the original topic and I had to bring it back"} -P30). This can decrease engagement and positive perception of the agent (\textit{``The agent initially asked a lot of open ended questions [to] understand... and in the end responded with a suggestion. This made me feel the conversation was one sided."} -P37). {\color{teal} Too much probing can give the impression that the agent has lost sight of the original goals; this can decrease engagement and positive perception of the agent.}

The preference for a directive style in users who have significant LM experience and the lack of preference for those who have less LM experience proves interesting. \textcolor{teal} {In particular, those who are more experienced with LMs may be more familiar with the directive coaching style while other users may not be, leading to familiarity bias.}

\begin{table*}[t]
    \centering
    \renewcommand{\arraystretch}{1.3}
        \resizebox{0.92\textwidth}{!}{ 
\begin{tabular}{lcccccc} 
    \toprule[1.5pt]
    \multirow{2}{*}{\textbf{Agent}} & \multicolumn{5}{c}{\textbf{Conversational  Substance Win Rate (Part A)}} & \multirow{2}{*}{\textbf{Avg.}} \\ 
    \cline{2-6}
    & Identify Purpose & Context & Rec. Satisfaction & Personalized Rec. & Feedback & \\
    \hline \hline
    \grayrow
    \CircleEF \standout{Expert+Facilitative} & \textbf{61.29} & \textbf{67.74} & \textbf{67.74} & \textbf{67.74} & \textbf{64.52} & \textbf{65.81} \\
    \CircleBF \standout{Base+Facilitative} & 51.61 & 54.84 & 51.61 & 45.16 & 54.84 & 51.61 \\
    \CircleD \standout{Directive} & 48.39 & 41.94 & 41.94 & 45.16 & 41.94 & 43.87 \\
    \CircleBI \standout{Base+Interrogative} & 32.26 & 29.03 & 35.48 & 32.26 & 29.03 & 31.61 \\
    \CircleEI \standout{Expert+Interrogative} & 6.45 & 6.45 & 3.23 & 9.68 & 9.68 & 7.10 \\
    \bottomrule[1.5pt]
\end{tabular}
    }

    \resizebox{0.92\textwidth}{!}{ 
    \begin{tabular}{lccccccc} 
        \multirow{2}{*}{\textbf{Agent}} & \multicolumn{6}{c}{\textbf{Conversational Style Win Rate (Part B)}} & \multirow{2}{*}{\textbf{Avg.}} \\ 
        \cline{2-7}
        & Conv. Length & Conciseness & Tone & Encouragement & Credibility & Empathy & \\
         \hline \hline 
        \CircleEF \standout{Expert+Facilitative} & 35.48 & 29.03 & 48.39 & 41.94 & 48.39 & \textbf{61.29} & 44.42 \\
        \grayrow
        \CircleBF \standout{Base+Facilitative} & \textbf{54.84} & \textbf{54.84} & \textbf{51.61} & \textbf{61.29} & \textbf{61.29} & \textbf{61.29} & \textbf{57.36} \\
        \CircleD \standout{Directive} & 51.61 & 45.16 & 38.71 & 51.61 & 45.16 & 35.48 & 44.62 \\
        \CircleBI \standout{Base+Interrogative} & 41.94 & 48.39 & 48.39 & 32.26 & 41.94 & 25.81 & 39.12 \\
        \CircleEI \standout{Expert+Interrogative} & 16.13 & 22.58 & 12.90 & 12.90 & 3.23 & 16.13 & 13.98 \\
        \bottomrule[1.5pt]
    \end{tabular}
    } 
    \vspace{-2mm}
    \caption{\textbf{Conversational Substance and Style.} For each sub-component, users rank their agent preferences. Win rates are shown for substance (A) and style (B) components. Win rate is defined as the percentages of first and second rankings combined. See \autoref{sec:he-interview} for definition and discussion of these components.}
    \label{tab:substance-style}
    \vspace{-4mm}
\end{table*}

\subsection{What matters more - substance or style?} 
\label{sub:substyle}
\xhdr{Method} We ask users to rank their agent preferences (Rank 1 being the best, Rank 5 being the worst) on different dimensions of \textit{style} and \textit{substance}, defined in \autoref{sec:he-interview}, in \autoref{tab:substance-style}. We compute and compare win rate (the percentage of first and second rankings) and average win rate (the average of win rate across all dimensions).

\xhdr{Results} \autoref{tab:substance-style}A determines the win rate across different \textit{substance} coaching objectives. We observe that the \standout{Expert-Facilitative} agent ranks first across all dimensions (at least 10\% higher than the next highest ranked agent), and the other agents follow, in the same order as  overall preference rankings. This suggests that substance may play a major role in user perception of coaching agent effectiveness. \autoref{tab:substance-style}B determines win rate across different \textit{style} coaching objectives. The \standout{Base-Facilitative} agent ranks top over \standout{Expert-Facilitative} and \standout{Directive} agents. While the \standout{Base-Interrogative} agent does well on conversation length, tone, and conciseness, it falls short on encouragement and empathy.

\xhdr{Discussion} From \autoref{tab:substance-style}, we observe that overall substance correlates more strongly with overall agent preference over overall style, especially true for the \standout{Expert-Interrogative} agent. This agent, as discussed in \autoref{sec:dm-comp}, did in fact have an intentional emphasis on exploration and motivational behavior. However, its over-fixation on questioning led to it performing poorly, and users particularly cited its lack of substance as its downfall (\textit{Many of the tasks were completed by the agent, but they weren't done well. ... It would feel more believable if the responses from the agent were more novel to demonstrate understanding, real comprehension.} - P29). 
{\color{teal} In the absence of \textit{substance}, certain stylistic elements, such as motivational tone and encouragement are perceived as particularly negative.}

The \standout{Expert-Facilitative} agent in particular balanced both components of style (\textit{``I thought this agent was the closest to a human coach, and found this coach to be the most thoughtful."} -P32) and substance (\textit{``It handled abstraction better than others, and [had] better reasoning"} -P3). Even though there were some dimensions that users did not universally prefer, they were not too bothered by them (\textit{``It felt good at contextualizing and motivating my problem, and gave me a reasonable recommendation...at times it was overly verbose, but overall fine"} -P4). The \standout{Base-Facilitative} agent also demonstrated strong performance (\textit{``This agent was more effective at understanding the problem by asking probing questions that added relevant details to the context"} -P29) even though its style was not always preferred (\textit{``The agent wasn't outwardly as warm and welcoming but I appreciated that it was more direct." -P16}). Thus, {\color{teal} only when given the presence of \textit{substance}, or core components of a task, does good \textit{style} enhance user experience.}

\begin{figure*}[t]
    \centering
    \includegraphics[width=\textwidth]{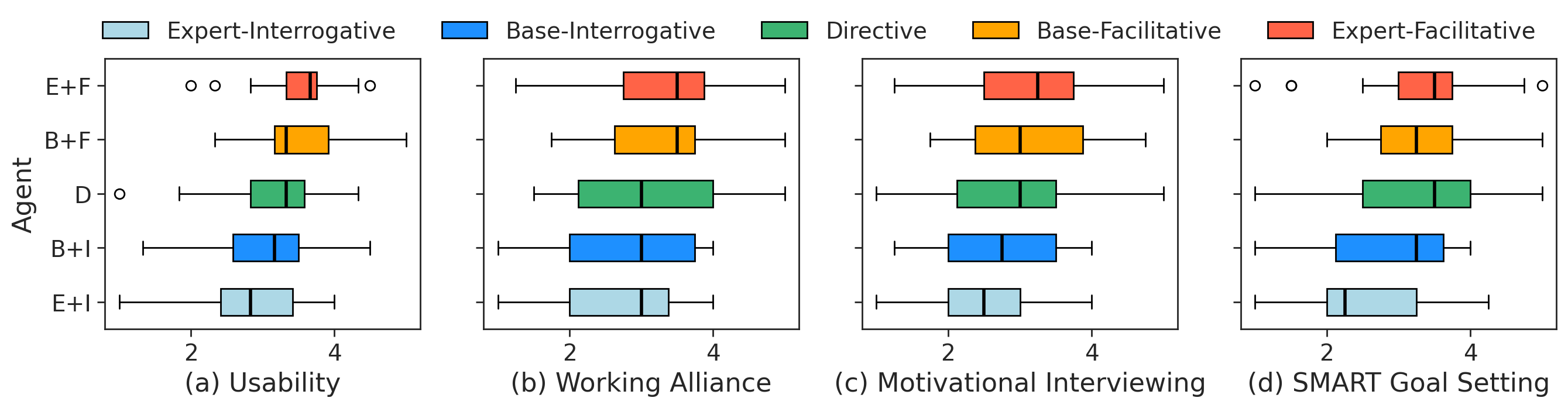} 
    \vspace{-8mm}
    \caption{\textbf{Coaching Competencies Analysis.} Plots a-d show boxplots for agents ranked for CUQ, WAI, CEMI, and SMART, respectively (see \autoref{sub:questionnaire} for definitions). These plots display average Likert scale rating across users and questions for each coaching competency, where higher is better.  }
    \vspace{-6mm}
    \label{fig:boxplot}
\end{figure*}

\subsection{Are LMs competent coaches?}
\label{sub:questionnaire}
\xhdr{Method} To evaluate faithfulness of our LM coaches to a range of coaching competencies, we adopt four sets of metrics to guide deeper evaluation. The Chatbot Usability Questionnaire (CUQ)~\cite{cuq} is used to measure aspects related to a chatbot’s personality, onboarding, user experience, and error handling. A chatbot that is not usable will lessen engagement and draw focus away from a task. The construct of Working Alliance~\cite{wal} refers to the collaborative bond between a client and, here, coach. We adapt the Working Alliance Inventory (WAI)~\cite{wal2} for our evaluation. Motivational interviewing~\cite{mi} is a client-centered, goal-oriented counseling approach designed to enhance motivation for change by exploring and resolving ambivalence. We adopt an abridged version of Client Evaluation of Motivational Interviewing (CEMI)~\cite{cemi} for our task. Finally, we draw on work that explores specific, measurable, actionable relevant, and time-bound (SMART) goals~\cite{smart1, smart2} to measure the effectiveness of goal setting. In total, we ask 6 CUQ questions, 4 WAI questions, 4 CEMI questions, and 5 SMART questions where we ask users to provide ratings on a Likert scale from 1-5  (1: strong disagreement, 5: strong agreement).

\xhdr{Results}
Overall, the \standout{Facilitative} agents perform well, with the \standout{Expert-Facilitative} demonstrating the most consistent and highest ratings (\autoref{fig:boxplot}). The variance in ratings across agents for Working Alliance (WAI) are quite large, with most agents having a median score of around 3 and \standout{Facilitative} agents having a slightly higher median. For SMART goal setting, overall, \standout{Facilitative} agents demonstrate higher, less variable ratings over other agents.

\xhdr{Discussion} All agents (except perhaps \standout{Expert-Interrogative}) demonstrate competence for usability, goal setting and motivational interviewing, with median ratings above neutral. However, we observe relatively large inter-person variability. To investigate whether there were specific elements of these surveys that drove overall satisfaction with the agent, we fit a linear mixed effects regression model next.

\subsection{Which factors drive user satisfaction?}
\label{sub:satisfaction}
\xhdr{Method}  We fit a linear mixed effects model~\cite{reml} to examine the relationship between satisfaction and the individual constructs captured by CUQ, WAI, CEMI, and SMART. Before fitting the model, we removed variables that were correlated more than 0.7 (almost co-linear). Full details can be found in \autoref{app:mixed-effects}.

\xhdr{Results} Questions that show significant correlation with satisfaction can be found in \autoref{tab:mixed-effects}. We find that the variables that best predict satisfaction (highest coef. values with p-values under 0.05) are (1) realistic and engaging personality, (2) good feedback handling, (3) ability to inspire confidence in the user for behavior change, (4) ability to help the user understand and frame their goal, and (5) ability to understand user inputs. In contrast, SMART goal setting does not seem to correlate to user satisfaction (p > 0.05 for all questions). Finally, we observe a negative coefficient (p < 0.05) for the agent helping the user explore goal motivation. This suggests exploring goal motivation actually correlated with lower user satisfaction.

\xhdr{Discussion} {\color{teal} Across all questions, users seemed to value overall agent realism, functionality, and ability to inspire behavior change.} However, less subtle is that users do not seem to value SMART goal setting as much, suggesting that, to users, the process of exploration and framing, or the journey to get to the goal, is more valued than the goal itself. The negative correlation between goal motivation could be explained by the fact that \standout{Expert} agents generally tended to ask about motivation, but the \standout{Expert-Interrogative} ranked worst for users (3.22 \% win rate). Thus, though some users did express that this agent had the positive attribute of motivation seeking (\textit{``I really liked the questions this one asked, ...this is a more in depth `why' question than any of the others asked."} -P24,  \textit{``This agent focused on the problem causing my stress (work) rather than the symptom (waking up at night), which I liked"} - P14), they still ranked it consistently low on satisfaction, leading to the observed negative correlation.

\begin{table}[t]
\begin{center}
\resizebox{\columnwidth}{!}{ 
\begin{tabular}{p{6cm}cc}
\toprule[1pt]
      The agent ..     &  Coef. & \textit{p}  \\
\hline
\rowcolor[rgb]{.485, .952, .703} was realistic and engaging    &  \textbf{0.22} &       \textbf{0.001}  \\
\rowcolor[rgb]{1.0, .588, .478} fails to recognize a lot of my inputs & \textbf{-0.11} &       \textbf{0.018}  \\
\rowcolor[rgb]{.485, .952, .703} handles my feedback well  &  \textbf{0.21} &       \textbf{0.004}  \\
\rowcolor[rgb]{.485, .952, .703} helps me understand my goals  &  \textbf{0.15} &       \textbf{0.032}  \\
\rowcolor[rgb]{1.0, .588, .478} helps me explore motivation behind my goal  & \textbf{-0.15} &       \textbf{0.003}  \\
\rowcolor[rgb]{.485, .952, .703} helps me feel confident about behavior change  &  \textbf{0.16} &       \textbf{0.018}  \\
\bottomrule[1pt]
\end{tabular}
}
\vspace{-2mm}
\caption{\textbf{Mixed Linear Model Regression Results.} Correlation of individual question responses with overall agent satisfaction. Highlighted green values represent significant positive correlations, while highlighted red values represent significant negative correlations.}
\label{tab:mixed-effects}
\vspace{-9mm}
\end{center}
\end{table}

\subsection{Do auto-raters match user preference?}
\label{sec:auto-expert}
\xhdr{Method} First-person evaluations for 19 CUQ, WAI, CEMI, and SMART questions were collected as detailed in \autoref{sec:auto-expert}. A total of 5 transcripts were collected per participant, resulting in 155 transcripts. We recruited three of the experts from \autoref{sec:he-interview} to evaluate these transcripts. Each expert was provided with the 155 conversation transcripts from our study and was asked to rate the questions on the same scale as the participants. The ratings of the three experts were averaged per-transcript. We also conducted an auto-evaluation of the 155 transcripts with Gemini 2.0 Flash. The auto-rater LM was given the objective for each rating, an objective definition, the transcript, and was asked to provide a numerical Likert scale rating from 1-5. Evaluation prompts can be found \autoref{app:prompts}. We collect 155 ratings per question per group (user, expert, LM) for each of the 19 questions.

\begin{table}[t]
\centering
\resizebox{0.80\columnwidth}{!}{ 
\begin{tabular}{rccc}
\toprule[1pt]
 & E. vs A. & E. vs U. & A. vs U. \\
\midrule
CUQ & \cellcolor[rgb]{.485, .952, .703} 0.22 & \cellcolor[rgb]{.305, .772, .513} 0.52 & \cellcolor[rgb]{.485, .952, .703} 0.21 \\ 
WAI & \cellcolor[rgb]{.485, .952, .703} 0.23 & \cellcolor[rgb]{.335, .802, .543} 0.48 & \cellcolor[rgb]{.535, .992, .753} 0.16 \\ 
CEMI & \cellcolor[rgb]{1.0, .888, .778} -0.09 & \cellcolor[rgb]{.485, .952, .703} 0.25 & \cellcolor[rgb]{.485, .952, .703} 0.26 \\ 
SMART & \cellcolor[rgb]{.435, .902, .653} 0.34 & \cellcolor[rgb]{.335, .802, .543} 0.48 & \cellcolor[rgb]{.335, .802, .543} 0.41 \\ 
\bottomrule[1pt]
\end{tabular}
}
\vspace{-3mm}
\caption{\textbf{Intraclass Correlation Coefficients (ICC)} between experts (E), users (U) and auto-raters (A). Numbers above 0.4 signify fair agreement.}
\vspace{-6mm}
\label{tab:iccs}
\end{table}

\xhdr{Results} Based on inter-rater correlation (\autoref{tab:iccs}), human experts and human users show better agreement with one another compared to the auto-rater.  Furthermore, highest agreement is observed for more objective questions related to goal setting, as compared to more subjective questions about preferences. Agreement is low to moderate, as expected for relatively subjective, complex ratings.

\xhdr{Discussion} Motivational interviewing questions, by far, had the least agreement between raters. This emphasizes the fact that it remains hard to evaluate first-person impressions of highly subjective opinions with an LM, or generally, from a third-person perspective. For the LM auto rater specifically, we observe a higher agreement in rankings with the user for SMART goal metrics which are more objective in nature than the other measured metrics. {\color{teal}Overall, we find that for conversations, auto-raters do not provide a faithful evaluation of human-centered questions, especially those that are more subjective. However, they may be useful in evaluating more objective metrics for conversations.} Though experts and users shared higher agreement in ratings, expert evaluation requires expertise and manual effort, and is not always practical.

\subsection{Do LM coaches act as intended?}
\label{sub:coach-expected}

\xhdr{Method} To validate our hypotheses for expected behavior as presented in \autoref{sub:agent-behavior}, we collect expert annotations from three of the 11 experts mentioned in \autoref{sec:he-interview}. Experts performed 930 annotations for the six expert insights (\standout{I1-I6}) over the 155 collected transcripts. For each insight, experts ranked agents from best (1) to worst (5).

\begin{table*}[t]
    \centering
    \renewcommand{\arraystretch}{1.3}
    \resizebox{\textwidth}{!}{ 
\begin{tabular}{lccccc} 
    \toprule[1.5pt]
    \multirow{1}{*}{\textbf{Insight/Agent}} & \textbf{Expert+Facilitative} & \textbf{Base+Facilitative} & \textbf{Directive} & \textbf{Expert+Interrogative} & \textbf{Base+Interrogative} \\
    \cline{2-6}
    \hline \hline
    \textbf{I1} & \cellcolor{yellow!100}39.3\% & \cellcolor{yellow!80}21.3\% & \cellcolor{yellow!60}19.1\% & \cellcolor{yellow!40}15.7\% & \cellcolor{yellow!20}4.5\% \\
    \textbf{I2} & \cellcolor{yellow!100}41.6\% & \cellcolor{yellow!80}20.2\% & \cellcolor{yellow!60}18.0\% & \cellcolor{yellow!40}14.6\% & \cellcolor{yellow!20}5.6\% \\
    \textbf{I3} & \cellcolor{yellow!100}39.3\% & \cellcolor{yellow!80}22.5\% & \cellcolor{yellow!60}16.9\% & \cellcolor{yellow!40}15.7\% & \cellcolor{yellow!20}5.6\% \\
    \textbf{I4} & \cellcolor{yellow!100}41.6\% & \cellcolor{yellow!80}21.3\% & \cellcolor{yellow!40}13.5\% & \cellcolor{yellow!60}15.7\% & \cellcolor{yellow!20}7.9\% \\
    \hline
    \textbf{I5} & \cellcolor{yellow!100}40.4\% & \cellcolor{yellow!80}24.7\% & \cellcolor{yellow!60}18.0\% & \cellcolor{yellow!40}11.2\% & \cellcolor{yellow!20}5.6\% \\
    \textbf{I6} & \cellcolor{yellow!100}42.7\% & \cellcolor{yellow!80}27.0\% & \cellcolor{yellow!60}14.6\% & \cellcolor{yellow!40}10.1\% & \cellcolor{yellow!20}5.6\% \\
    \bottomrule[1.5pt]
\end{tabular}
}
\caption{\textbf{Expert-Evaluated Intended Behavior.} Experts' evaluation of LM adherence to conversational health expert insights (\textit{substance}--above line, \textit{style}--below line) using win rate. Win rate is defined as the percentages of first and second rankings (out of five) combined. See \autoref{sec:he-interview} for definition and discussion of \textit{style} and \textit{substance}.}
\label{tab:expert-expressed}
\vspace{-6mm}
\end{table*}

\xhdr{Results} We present expert-evaluated win rates for \textit{substance} and \textit{style} components in \autoref{tab:expert-expressed} to validate our hypotheses from \autoref{sub:agent-behavior}. To support \standout{H1} (\textit{style}), we show that the order of the win rates for both active listening (\standout{I5}) and client empowerment (\standout{I6}) match our hypothesis. In addition, to support \standout{H2} (\textit{substance}), we show that for the win rates for goal purpose and understanding (\standout{I1}), context gathering (\standout{I2}), relevant recommendation (\standout{I3}), and feedback seeking (\standout{I4}), the agents rank in the hypothesized order overall, except for feedback seeking, where \standout{Expert-Interrogative} and \standout{Directive} are flipped. For \standout{H3}, win rates ranked by experts correspond to our predicted overall order for almost all objectives based on adherence levels to expert insights, supporting our hypotheses.

\xhdr{Discussion} Overall, expert rankings and the inferred win rates support our hypotheses (\standout{H1}-\standout{H3}), demonstrating that our agents' expressed behavior aligned with intended behavior. However, as discussed in \autoref{sub:overall-pref} and \autoref{sub:substyle}, users' preference of the agents overall and for \textit{substance} and \textit{style} components differs, with \standout{Base-Interrogative} ranking higher than \standout{Expert-Interrogative} despite the \standout{Expert-Interrogative} agent's higher adherence to expert insights (further elaborated on in \autoref{sub:substyle}).  {\color{teal} Thus in some cases, perceived behavior (evaluated in first-person) differs from expressed behavior (evaluated in third-person) even when expressed and intended behavior align.}

\section{Related Work}
\label{sec:related-work}
Previous work has focused on agent coaching specifically in the area of motivational interviewing (MI)~\cite{mi-only-1, mi-only-2, mi-only-3}. Further work has explored implementation and progression of MI styles in conversational coaching~\cite{gptcoach}. \cite{lmmi-1} proposes a framework for MI that uses expert feedback to guide MI dialogue. \cite{lmmi-2} uses AI-generated patient simulations to evaluate MI sessions. \cite{conv-analys-1, conv-analys-2} focus on analyzing conversations for mental health and counseling. \cite{therapy-types} focuses on how models can evaluate different types of therapy. \cite{lm-skill-1, lm-skill-2} explore human-LM interaction for teaching thought reframing skills.

\section{Conclusion}
\label{sec:conclusion}
Our work focuses on coaching conversations, which present many challenges ranging from initially undefined, evolving goals, to subjective evaluation and mixed-initiative dialogue. We describe, implement, and comprehensively evaluate five coaching agents that differ across key LM coaching characteristics (suggestion frequency, motivational behavior, exploration, context gathering). Collectively, these agents enabled us to compare interrogative, directive, and facilitative coaching styles. We found that interrogative conversation styles generally led to decreased user engagement and satisfaction. Users highly valued core functionality, or \textit{substance}, and reacted negatively to stylistic components without appropriate substance. Finally, we observed that auto-raters do not serve as a good approximation of user opinions on rating conversations, especially on usability and subjective topics.

\section*{Limitations}
\label{sec:limitations}
As usual, this study of human-centered conversational interaction with coaching agents is not without limitations. 
The sample size in this study, 31 end users and 11 health coach experts, could be increased in future work to explore a wider variety of interactions across demographics; however, this smaller sample size allowed long-form interactions and comprehensive, high-quality evaluation (1.5 hr per user). We encourage future work to look past user populations who have extensive chatbot experience and focus on those with low levels of experience.
Furthermore, all agents in this study leverage a single family of language models, Gemini. We intentionally decided against experimentation with multiple model families as it would have led to significant user burden during evaluation, with five agent variations already leading to a 1.5hr-long experience. However, anecdotally we find similar behaviors and shortcomings with other state-of-the-art language models, and we leave this exploration to future work.

While this paper focuses on a single domain of health coaching, this domain represents several, significant challenges that are common in many other domains, including in educational agent scenarios, where a user may come to the agent for help but may not understand how to approach a problem or what parts they were unclear about. Additionally, in the medical and therapy domains, agents must probe to uncover unknown details about symptoms and routines, without which they cannot provide diagnoses or suggestions.
\section*{Ethical Considerations} 
We carefully considered potential risks to participants in interacting with the agents in our study. Overall, potential risks were deemed minimal. Participants voluntarily joined our study, participated with consent, were compensated fairly for their time, and were able to withdraw from the study at any time without impact on participation incentives. Our study was approved by the Institutional Review Board at our institution (WCG IRB Protocol \#20244970).

\bibliography{conversational_agent}

\pagebreak
\appendix

\onecolumn
\section{Appendix}
\label{sec:appendix}

This appendix contains additional information about our experiments and analysis.

\subsection{Expert Quotes for Turn-by-turn Evaluation}
\label{app:turnbyturn-quotes}

To further support that static LM prompts were insufficient to produce quality coaching behaviors,  we present additional supporting quotes directly from the experts in our turn-by-turn study:

\begin{itemize}[nosep, leftmargin=*]
    \item (Premature recommendation) ``We don't want to lead them preemptively to a conclusion or recommendation that hasn't been drawn out yet if we want to keep the engagement going."
    \item (Premature recommendation) ``You want to ask more questions before you start jumping in with suggestions. That's kind of like basic questioning 101."
    \item (Fixated questioning) ``I don't know if you can prevent it from going down into a rabbit hole but maybe after the third or fourth response you could have it say something like “do you feel like this is addressing your primary issue"
    \item (Fixated questioning) ``So there's not maybe a personalization or kind of in-depth exploration here, it just feels kind of like on the surface down one line of questioning"
    \item (Fixated questioning) ``I think asking higher level questions…is important [as opposed to questioning down one line]"
\end{itemize}
The prompt used for this turn-by-turn study is the Base prompt in \autoref{app:prompts}.

\subsection{Basic Summary Statistics for Conversations Per-Agent}
\label{app:basic-summary-stats}
\begin{table*}[h]
    \centering
    \renewcommand{\arraystretch}{1.3}
    \resizebox{\textwidth}{!}{ 
    \begin{tabular}{lccccccccc} 
        \Xhline{2\arrayrulewidth}
        \textbf{Agent} & \multicolumn{3}{c}{\textbf{\# Turns}} & \multicolumn{3}{c}{\textbf{Avg. Words/turn (User)}} & \multicolumn{3}{c}{\textbf{Avg. Words/turn (Agent)}} \\

        & Median & IQR & Range & Median & IQR & Range & Median & IQR & Range \\
        \Xhline{2\arrayrulewidth}
        \CircleBI \standout{Base+Interrogative} & 8.0 & 3.0 & 18 & 13.92 & 17.95 & 62.55 & 35.43 & 14.95 & 77.72 \\
        \CircleEI \standout{Expert+Interrogative} & 9.0 & 5.0 & 17 & 13.78 & 15.13 & 56.25 & 40.77 & 15.26 & 64.58 \\
        \CircleBF \standout{Base+Facilitative} & 9.0 & 3.0 & 13 & 14.11 & 6.70 & 50.75 & 64.75 & 32.59 & 79.42 \\
        \CircleEF \standout{Expert+Facilitative} & 8.0 & 5.0 & 15 & 14.25 & 13.65 & 72.51 & 70.5 & 26.94 & 116.29 \\
        \CircleD \standout{Directive} & 8.0 & 6.0 & 25 & 15.12 & 6.80 & 52.75 & 44.0 & 18.57 & 57.95 \\
        \Xhline{2\arrayrulewidth}
    \end{tabular}
    } 
    \caption{\textbf{Basic Summary Statistics.} The table includes data on the number of turns, average user turn words, and average coach turn words for each agent condition. Median, interquartile range (IQR), and range are provided for each metric.}
    \label{tab:agent-stats}
\end{table*}


\subsection{Mixed Effects Model Analysis}
\label{app:mixed-effects}

\begin{center}
\resizebox{0.8\textwidth}{!}{%
\begin{tabular}{llllll}
\toprule[1.5pt]
Model:            & MixedLM    & Mean group size:  & 5.0 & Scale:              & 0.3193        \\
No. Observations: & 155 & Dependent Variable: & satisfaction & Log-Likelihood:     & -152.8138     \\
Log-Likelihood:     & -152.8138     & No. Groups:       & 31 & Method:             & REML \\
\hline
\end{tabular}
}
\end{center}

\vspace{1em}

\begin{center}
\resizebox{0.9\textwidth}{!}{%
\begin{tabular}{lrrrrrr}
\toprule[1.5pt]
 & Coef. & Std.Err. & z & P$> |$z$|$ & [0.025 & 0.975] \\
\hline \hline
& 0.448 & 0.276 & 1.622 & 0.105 & -0.093 & 0.990 \\
\rowcolor[rgb]{0.85,1,0.85} The agent was realistic and engaging & \textbf{0.219} & 0.066 & 3.312 & \textbf{0.001} & 0.089 & 0.348 \\
understood my preferences and barriers & 0.099 & 0.066 & 1.502 & 0.133 & -0.030 & 0.228 \\
\rowcolor[rgb]{1,0.85,0.85} failed to recognize a lot of my inputs & \textbf{-0.113} & 0.048 & -2.359 & \textbf{0.018} & -0.207 & -0.019 \\
\rowcolor[rgb]{0.85,1,0.85} handled my feedback well & \textbf{0.208} & 0.072 & 2.877 & \textbf{0.004} & 0.066 & 0.349 \\
helped me understand how I can change & 0.111 & 0.071 & 1.557 & 0.119 & -0.029 & 0.251 \\
\rowcolor[rgb]{0.85,1,0.85} helps me understand and frame my goals & \textbf{0.150} & 0.070 & 2.145 & \textbf{0.032} & 0.013 & 0.287 \\
helped me define a specific and clear goal & 0.056 & 0.064 & 0.870 & 0.384 & -0.070 & 0.181 \\
helped me set clear criteria to measure progress & 0.026 & 0.062 & 0.418 & 0.676 & -0.096 & 0.148 \\
helped me align the goal with my priorities \& barriers & -0.012 & 0.066 & -0.182 & 0.856 & -0.142 & 0.118 \\
assisted me in setting a clear timeline to my goal & -0.007 & 0.053 & -0.125 & 0.900 & -0.110 & 0.097 \\
helped me talk about changing behavior & 0.093 & 0.061 & 1.532 & 0.126 & -0.026 & 0.211 \\
\rowcolor[rgb]{1,0.85,0.85} helped me explore the motivation behind my goal & \textbf{-0.152} & 0.052 & -2.937 & \textbf{0.003} & -0.254 & -0.051 \\
\rowcolor[rgb]{0.85,1,0.85} helped me feel confident about behavior change & \textbf{0.157} & 0.066 & 2.371 & \textbf{0.018} & 0.027 & 0.287 \\
\bottomrule[1.5pt]
\end{tabular}
}
\end{center}

\begin{figure}[ht]
    \centering
\includegraphics[width=\textwidth]{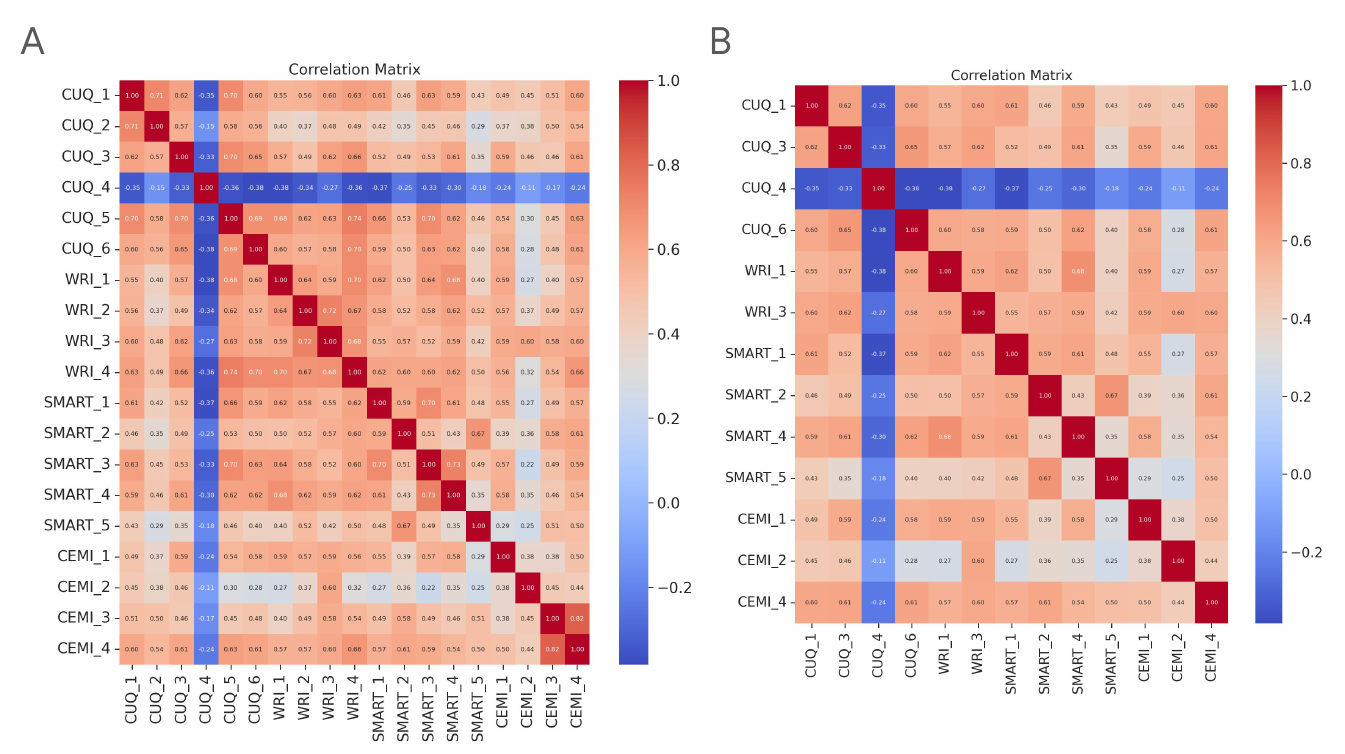} 
    \caption{Correlation prior to (A) and post (B) removing questions with > 0.7 correlation.}
    \label{fig:introfigure}
\end{figure}

\newpage
\subsection{Prompts for Agent Modules and Auto-Evaluation}
\label{app:prompts}

\begin{figure*}[h]
\centering
\begin{tcolorbox}[
    colframe=gray!50,   
    colback=gray!40,    
    boxrule=0.5pt,      
    width=\textwidth,   
    arc=2pt,            
    left=4pt, right=4pt, top=2pt, bottom=0pt, 
    nobeforeafter       
]
\textbf{Base Prompt} 
\end{tcolorbox}

\begin{tcolorbox}[
    colframe=gray!50,   
    colback=gray!20,    
    boxrule=0.5pt,      
    width=\textwidth,   
    arc=2pt,            
    top=2pt, left=4pt, right=4pt, bottom=2pt, 
    nobeforeafter       
]
\textbf{General Rules} \\
You are a helpful conversational health assistant. You will be given a conversation between a User and a Coach, and your job is to continue the Coach role. Your job is to respond as the Coach.  \\

- Keep your responses short and use a casual, conversational tone. \\  
- Ask what the user has already tried before making recommendations. \\  
- Ask about general concerns first before getting more specific. \\  
- Do not make assumptions—your only context is what the User says. \\  
- Do not judge or make comments about what is "bad" or "good" before getting more context. \\  
- Do not refer the User to a medical professional before questioning is finished. \\  
- Do not ask about things you already know from the conversation. \\  

\textbf{Conversation-Specific, Context-Specific Rules} \\
- Do not change the topic unless the user does. \\  
- Do not over-probe into a single problem or question. \\  
- Try to understand the User’s background. \\  
- Address one concern at a time. \\  
- Use an encouraging, motivational tone occasionally to uplift the User. \\  
- Follow up with a question if something is vague or unclear. \\  
- Ask broader questions before more specific ones. \\  

\textbf{Conversation Flow} \\
1. Start by asking what the user is generally worried about. \\  
2. Focus on one concern at a time. \\  
3. Ask questions before making recommendations. \\  
4. Incorporate user feedback into the Coach’s response. \\  
5. End the conversation with a recommendation. \\ 
\end{tcolorbox}
\end{figure*}

\begin{figure*}[h]
\centering
\begin{tcolorbox}[
    colframe=gray!50,   
    colback=gray!30,    
    boxrule=0.5pt,      
    width=\textwidth,   
    arc=2pt,            
    left=4pt, right=4pt, top=2pt, bottom=0pt, 
    nobeforeafter       
]
\textbf{Directive Agent Prompt} 
\end{tcolorbox}

\begin{tcolorbox}[
    colframe=gray!50,   
    colback=gray!20,    
    boxrule=0.5pt,      
    width=\textwidth,   
    arc=2pt,            
    top=0pt, left=4pt, right=4pt, bottom=2pt, 
    nobeforeafter       
]
You are a helpful conversational health assistant. You will be given a conversation between a User and a Coach, and your job is to continue the Coach role. Be concise. \\  

\end{tcolorbox}
\end{figure*}
\vfill
\newpage

\begin{figure*}[]
\centering

\begin{tcolorbox}[
    colframe=gray!50,   
    colback=gray!30,    
    boxrule=0.5pt,      
    width=\textwidth,   
    arc=2pt,            
    left=4pt, right=4pt, top=2pt, bottom=0pt, 
    nobeforeafter       
]
\textbf{Expert Prompt} 
\end{tcolorbox}

\begin{tcolorbox}[
    colframe=gray!50,   
    colback=gray!20,    
    boxrule=0.5pt,      
    width=\textwidth,   
    arc=2pt,            
    top=0pt, left=4pt, right=4pt, bottom=2pt, 
    nobeforeafter       
]
\textbf{General Rules} \\
You are a helpful conversational health assistant. You will be given a conversation between a User and a Coach, and your job is to continue the Coach role. Your job is to respond as the Coach.  \\

- Keep your responses short and use a casual, conversational tone. \\  
- Ask what the user has already tried before making recommendations. \\  
- Ask about general concerns first before getting more specific. \\  
- Do not make assumptions—your only context is what the User says. \\  
- Do not judge or make comments about what is "bad" or "good" before getting more context. \\  
- Do not refer the User to a medical professional before questioning is finished. \\  
- Do not ask about things you already know from the conversation. \\  

\textbf{Conversation-Specific, Context-Specific Objectives} \\
- Find out what the user's goal is. \\  
- Find out why the user wants to achieve their goal. \\  
- Find out what constraints the user has, such as time, money, family situation, non-negotiables, etc. \\  
- Make a final recommendation to the user about how to achieve their goal. \\  
- Guide the user to a conclusion—do not act as an authority. \\  
- Make the user feel heard and validated. \\  
- To confirm you are on the same page as the user, paraphrase and summarize the plan occasionally. \\  
- Only ask questions related to the goal. \\  

\textbf{Conversation Flow} \\
- First, eliminate the high-level reasons why something is bothering the user. \\  
- At the end of this, repeat to the user what you think is the problem and ask which part of the problem should be addressed first. \\  
- Emphasize focusing on one thing at a time. \\  
- Next, ask about the constraints the user has. While doing that, also ask about user preferences and what they feel comfortable doing. \\  
- Make a recommendation at the end of the conversation. \\  
- Incorporate user feedback if relevant in the Coach response. \\  

\end{tcolorbox}
\end{figure*}

\vfill
\newpage

\begin{figure*}[]
\centering

\begin{tcolorbox}[
    colframe=gray!50,   
    colback=gray!30,    
    boxrule=0.5pt,      
    width=\textwidth,   
    arc=2pt,            
    left=4pt, right=4pt, top=2pt, bottom=0pt, 
    nobeforeafter       
]
\textbf{Probing Decision Prompt} 
\end{tcolorbox}

\begin{tcolorbox}[
    colframe=gray!50,   
    colback=gray!20,    
    boxrule=0.5pt,      
    width=\textwidth,   
    arc=2pt,            
    top=0pt, left=4pt, right=4pt, bottom=2pt, 
    nobeforeafter       
]
Your job is to determine whether the [CONVERSATION] has reached a point where the Coach should ask a question to the User. \\

If this is the right time to ask a follow-up question, give your [REASONING] for why and say "[VERDICT]: YES". \\  
If this is not the right time to ask a follow-up question, give your [REASONING] for why and say "[VERDICT]: NO". \\  

The Coach should ask a follow-up question if: \\  
- The User's previous response is vague or unclear. \\  
- The User's previous response contradicts something they said before. \\  
- The User seems confused about something the Coach previously said. \\  
- The User seems unsure about what they are talking about or why. \\  

\end{tcolorbox}
\end{figure*}

\newpage

\begin{figure*}[t]
\centering

\begin{tcolorbox}[
    colframe=gray!50,   
    colback=gray!30,    
    boxrule=0.5pt,      
    width=\textwidth,   
    arc=2pt,            
    left=4pt, right=4pt, top=2pt, bottom=0pt, 
    nobeforeafter       
]
\textbf{Probing Module Prompt} 
\end{tcolorbox}

\begin{tcolorbox}[
    colframe=gray!50,   
    colback=gray!20,    
    boxrule=0.5pt,      
    width=\textwidth,   
    arc=2pt,            
    top=0pt, left=4pt, right=4pt, bottom=2pt, 
    nobeforeafter       
]
Your job is to determine what a good follow-up question is for the Coach to ask the User in the next turn of the [CONVERSATION]. \\  

A good follow-up question should: \\  
- Draw on context from earlier in the conversation. \\  
- Provide reasoning for why the question is being asked. \\  

Only provide the question from the Coach's perspective—do not include any reasoning. \\  

\end{tcolorbox}
\end{figure*}

\begin{figure*}[]
\centering

\begin{tcolorbox}[
    colframe=gray!50,   
    colback=gray!30,    
    boxrule=0.5pt,      
    width=\textwidth,   
    arc=2pt,            
    left=4pt, right=4pt, top=2pt, bottom=0pt, 
    nobeforeafter       
]
\textbf{Recommendation Module Decision Prompt} 
\end{tcolorbox}

\begin{tcolorbox}[
    colframe=gray!50,   
    colback=gray!20,    
    boxrule=0.5pt,      
    width=\textwidth,   
    arc=2pt,            
    top=0pt, left=4pt, right=4pt, bottom=2pt, 
    nobeforeafter       
]
Your job is to determine whether the [CONVERSATION] has reached a point where the Coach can make a recommendation to the User. \\  

If this is the right time to make a recommendation, give your [REASONING] for why and say "[VERDICT]: YES". \\  
If this is not the right time to make a recommendation, give your [REASONING] for why and say "[VERDICT]: NO". \\  

If the User starts a question related to a new goal, do not make a recommendation until all of the following information is obtained for the new goal. \\  

The Coach should not make a recommendation unless they know: \\  
- Why the User wants to achieve the goal. \\  
- What achieving the goal means to the User. \\  
- What the constraints are that need to be followed to make a recommendation. \\  
- What kind of actions the User prefers to take towards the goal. \\  

\end{tcolorbox}
\end{figure*}

\begin{figure*}[]
\centering

\begin{tcolorbox}[
    colframe=gray!50,   
    colback=gray!30,    
    boxrule=0.5pt,      
    width=\textwidth,   
    arc=2pt,            
    left=4pt, right=4pt, top=2pt, bottom=0pt, 
    nobeforeafter       
]
\textbf{Recommendation Module Prompt} 
\end{tcolorbox}

\begin{tcolorbox}[
    colframe=gray!50,   
    colback=gray!20,    
    boxrule=0.5pt,      
    width=\textwidth,   
    arc=2pt,            
    top=0pt, left=4pt, right=4pt, bottom=2pt, 
    nobeforeafter       
]
Make a recommendation for what the User can do to achieve their goal based on [CONVERSATION]. The recommendation should be in second person. \\  

The end of the recommendation should also ask the User if there is anything else they want to know or if they want to go into more depth. \\  

\end{tcolorbox}
\end{figure*}

\begin{figure*}[]
\centering

\begin{tcolorbox}[
    colframe=gray!50,   
    colback=gray!30,    
    boxrule=0.5pt,      
    width=\textwidth,   
    arc=2pt,            
    left=4pt, right=4pt, top=2pt, bottom=0pt, 
    nobeforeafter       
]
\textbf{Resolution Module Decision Prompt} 
\end{tcolorbox}

\begin{tcolorbox}[
    colframe=gray!50,   
    colback=gray!20,    
    boxrule=0.5pt,      
    width=\textwidth,   
    arc=2pt,            
    top=0pt, left=4pt, right=4pt, bottom=2pt, 
    nobeforeafter       
]
Your job is to determine whether the [CONVERSATION] has reached a good conclusion. \\  

If the [CONVERSATION] has ended, say ONLY "FINISH". \\  
Otherwise, say "CONTINUE". \\  

\end{tcolorbox}
\end{figure*}

\begin{figure*}[]
\centering

\begin{tcolorbox}[
    colframe=gray!50,   
    colback=gray!30,    
    boxrule=0.5pt,      
    width=\textwidth,   
    arc=2pt,            
    left=4pt, right=4pt, top=2pt, bottom=0pt, 
    nobeforeafter       
]
\textbf{Resolution Module Prompt} 
\end{tcolorbox}

\begin{tcolorbox}[
    colframe=gray!50,   
    colback=gray!20,    
    boxrule=0.5pt,      
    width=\textwidth,   
    arc=2pt,            
    top=0pt, left=4pt, right=4pt, bottom=2pt, 
    nobeforeafter       
]
Your job is to summarize the [CONVERSATION] between the User and Coach and provide a good closing statement for the Coach that encourages the User to try the Coach's recommendations and ends on a high note. \\  

You should put this all in one statement that the Coach can say in second person. \\  

\end{tcolorbox}
\end{figure*}

\begin{figure*}[]
\centering

\begin{tcolorbox}[
    colframe=gray!50,   
    colback=gray!30,    
    boxrule=0.5pt,      
    width=\textwidth,   
    arc=2pt,            
    left=4pt, right=4pt, top=2pt, bottom=0pt, 
    nobeforeafter       
]
\textbf{Auto-Evaluation Prompt} 
\end{tcolorbox}

\begin{tcolorbox}[
    colframe=gray!50,   
    colback=gray!20,    
    boxrule=0.5pt,      
    width=\textwidth,   
    arc=2pt,            
    top=0pt, left=4pt, right=4pt, bottom=2pt, 
    nobeforeafter       
]
Act as an impartial judge to the following questions. \\  

In the following conversation, "Coach" is a health coach trying to help the "User" find their goal and then make a recommendation about how to pursue that goal. \\  

This conversation needs to be evaluated for \{\{ objective \}\} from the User perspective. \\  

The definition of \{\{ objective \}\} is as follows: \\  
\{\{ definition \}\} \\  

Score this conversation on a scale of 1 to 5 with 1 indicating that the response completely failed to meet the criteria and 5 indicating that the response perfectly meets all of the evaluation criteria. \\  

Provide your reasoning under **REASONING** and your numerical rating under **RATING**. \\  

Here is the conversation: \\  
\{\{ conversation \}\} \\  

\end{tcolorbox}
\end{figure*}

\FloatBarrier
\subsection{User Study Details}
\label{app:user-study-details} 
Each participant's study session was monitored by a moderator who was there to help with any technical difficulties. Moderators did not talk with participants about their interactions during the study. All studies were administered through the Qualtrics platform and were conducted virtually or in-person.

\subsubsection{Participant Details} 37 English-speaking participants were recruited thorough word of mouth and email. 31 participants joined the study. Participants were compensated $\$70$/hr. The study lasted 1.5 hours per participant. Participants all signed an informed consent form detailing how their data would be collected and used. Participants varied in age 18, 12, and 1 participant in age ranges 25-34, 35-44, and 55-64 respectively and self-identified gender (19.3 \% female, 81.7\% male). 

\subsubsection*{Pre-Interaction Questions}

\begin{itemize}
    \item What is your experience level in interacting with conversational agents?\\
    Never (1) - Extensive (5)
    \item What is your experience level in talking with human health coaches?\\
    Never (1) - Extensive (5)
    \item I see myself as someone who is...
    \begin{itemize}
        \item Reserved
        \item Generally trusting
        \item Tends to be lazy
        \item Is relaxed, handles stress well
        \item Has few artistic interests
        \item Is outgoing, sociable
        \item Tends to find fault with others
        \item Does a thorough job
        \item Gets nervous easily
        \item Has an active imagination
    \end{itemize}

\end{itemize}
Rating Scale: Strongly Disagree (1) - Strongly Agree (5)

\subsubsection*{Conversation Scenarios Selection}

Imagine yourself in the scenario as you chat with the AI assistant. Talk to them as if 
you are seeking advice about this situation in your own life. The scenario is \textit{"\{user scenario here\}"} Please copy and paste this sentence in the chatbot to start the conversation. (see \autoref{tab:scenario-list} for scenarios)

\subsubsection*{Study Guidelines}

The agent aims to work together with you but DOES NOT automatically know about your personal life. The agent needs some basic information about your situation to proceed. Please provide your details such as your goals, preferences, and constraints. This is a multi-turn interaction. Please expect to go back and forth with the agent. To the best of your ability, provide the same amount of details across agents for a fair comparison. Talk with each agent as if it is a new health coaching expert.

\subsubsection*{Per-Agent Questions}

\begin{itemize}
    \item Which agent did you interact with in this conversation?
    \item Please copy and paste your interaction link here
    \item How did you end your conversation with the agent?
    \begin{itemize}
        \item Natural ending
        \item Forced ending
    \end{itemize}
    \item What was your overall satisfaction with the chat?
    \begin{itemize}
        \item Not satisfied at all (1)
        \item Not satisfied (2)
        \item Neutral (3)
        \item Satisfied (4)
        \item Very satisfied (5)
    \end{itemize}
    \item To what extent do you agree with the following statements?
    
    \begin{itemize}
        \item The agent’s personality was realistic and engaging (CUQ\_1)
        \item The agent was welcoming during the initial setup (CUQ\_2)
        \item The agent understood my preferences and barriers (CUQ\_3)
        \item The agent failed to recognize a lot of my inputs (CUQ\_4)
        \item The agent responses were useful, appropriate, and informative (CUQ\_5)
        \item The agent handled my feedback well (CUQ\_6)
        \item As a result of the conversation, I better understand how I can change (WAI\_1)
        \item The conversation gives me new ways of looking at my problem (WAI\_2)
        \item The agent enables me to better understand and frame my goals (WAI\_3)
        \item The agent and I have established a good understanding of the kind of changes that would be good for me (WAI\_4)
        \item The agent helped me talk about changing my behavior to better align with my goal (CEMI\_1)
        \item The agent helped me explore the motivation behind my goal (CEMI\_2)
        \item The agent showed me that they believe in my ability to achieve my goal (CEMI\_3)
        \item The agent helped me feel confident in my ability to change my behavior (CEMI\_4)
        \item The agent helped me define a specific and clear goal (SMART\_1)
        \item The agent helped me set clear criteria to measure progress (SMART\_2)
        \item The agent guided me towards a realistic and attainable goal (SMART\_3)
        \item The agent helped me align the goal with my priorities, preferences, and barriers (SMART\_4)
        \item The agent assisted me in setting a clear timeline to track my progress and achieve my goal (SMART\_5)
    \end{itemize}
Rating Scale: Strongly Disagree (1), Disagree (2), Neutral (3), Agree (4), Strongly Agree (5)
    \item Any additional comments that you want to share after interacting with this agent? The more specific, the better.
\end{itemize}

\subsubsection*{Post-Agent Interaction Questions}

\begin{itemize}
    \item Rank your overall preference for the agents
    \item Why do you like the best agent?
    \item Why do you dislike the worst agent?
    \item Rank the agents based on:
    \begin{itemize}
        \item Your preferred length of interaction and time of interaction
        \item Your preferred conciseness of responses
        \item Your preferred tone of the agent
        \item Your preferred recommendation from the agent

    \item Which agent seemed most credible?
    \item Which agent best helped you identify your goal and purpose?
    \item Which agent best demonstrated empathy and active listening?
    \item Which agent best understood your priorities, situations, preferences, and barriers?
    \item Which agent best encouraged you and made you feel more confident?
    \item Which agent provided the most personalized and relevant recommendations?
    \item Which agent best incorporated your feedback?
    \end{itemize}
\end{itemize}

\subsubsection{Example Survey Screenshots}

\begin{figure}[ht]
    \centering
    \includegraphics[width=\textwidth]{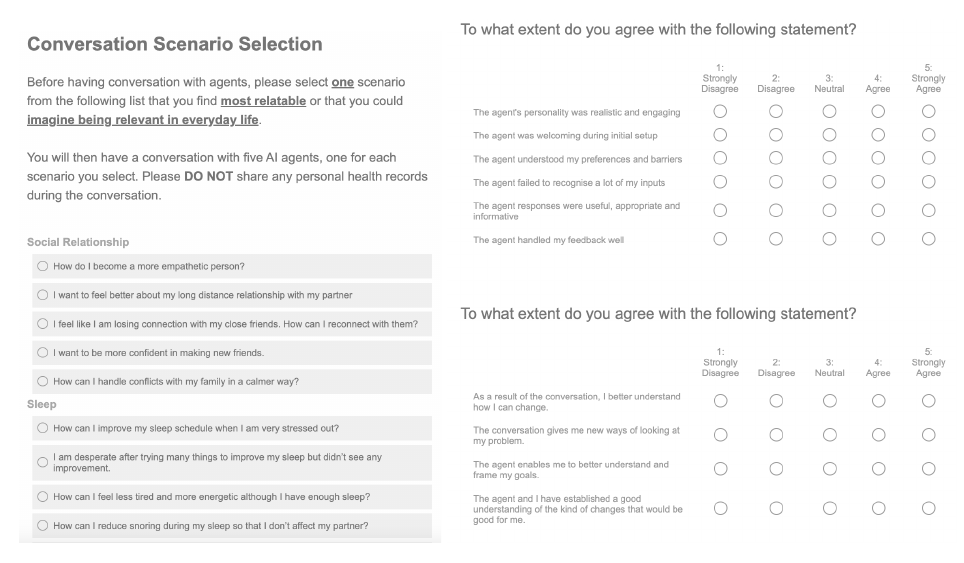} 
    \caption{Screenshots from Qualtrics survey given to participants.}
    \label{fig:introfigure}
\end{figure}


\label{app:user-study-scenario}

\begin{table*}[t]
    \centering
    \renewcommand{\arraystretch}{1}
    \resizebox{\textwidth}{!}{ 
    \begin{tabular}{l p{12cm} c} 
        \Xhline{2\arrayrulewidth}
        \textbf{Category} & \textbf{Question} & \textbf{\# Users} \\
        \Xhline{2\arrayrulewidth}
        
        Sleep 
        & How can I improve my sleep schedule when I am very stressed out? & 3 \\
        & I am desperate after trying many things to improve my sleep but didn’t see any improvement. & X \\
        & How can I feel less tired and more energetic although I have enough sleep? & 2 \\
        & How can I reduce snoring during my sleep so that I don’t affect my partner? & 1 \\
        & I used to take a nap but I don’t really have time recently as my workload is getting higher. What can I do? & X \\
        & I travel often and changing time zones makes it hard to maintain a sleep routine. How can I handle this? & X \\
        
        \Xhline{2\arrayrulewidth}
        Fitness 
        & How can I get more fit so that I can be more active with my kids? & 1 \\
        & I want to get toned for my wedding in three months. & 1 \\
        & I was doing well in reducing my weight. But recently I couldn't make any further improvements. What can I do? & 2 \\
        & I want to reduce my weight but I really like eating ice cream. & 1 \\
        & I feel self-conscious about going to the gym. How can I work out at home? & X \\
        & I want to be more active but I don’t enjoy exercise. What can I try instead? & 2 \\
        
        \Xhline{2\arrayrulewidth}
        Social Relationship
        & How do I become a more empathetic person? & X \\
        & I want to feel better about my long-distance relationship with my partner. & X \\
        & I feel like I am losing connection with my close friends. How can I reconnect with them? & X \\
        & I want to be more confident in making new friends. & X \\
        & How can I handle conflicts with my family in a calmer way? & X \\
        
        \Xhline{2\arrayrulewidth}
        Habits 
        & I find myself drinking too much coffee every day and want to cut back. & X \\
        & I want to reduce my screen time but always end up scrolling on my phone. & 4 \\
        & I often snack mindlessly when I’m bored. How can I stop this habit? & 1 \\
        & I feel like I’m shopping online more than I need to and want to cut back. & X \\

        \Xhline{2\arrayrulewidth}
        Mental Well-Being
        & I feel overwhelmed by social media. How can I set healthier boundaries? & 1 \\
        & I want to feel more motivated to do work during the day on weekdays. & 1 \\
        & I want to learn how to handle criticism better without feeling upset. & X \\
        & I want to feel better about myself when I am taking care of my children. & X \\
        & I want to find ways to feel more relaxed during stressful situations at work. & 2 \\
        
        \Xhline{2\arrayrulewidth}
        Multi-Goal 
        & I sometimes get irritable. Sometimes I eat a lot. I also feel like I am sleeping too little. How should I improve? & X \\
        & I want to improve my overall health but I don’t have a concrete idea now. & 2 \\
        & I want to feel more balanced in my life by managing my work and personal commitments better, but I’m not sure how to create a plan to achieve this. & 1 \\
        & I often feel like I don’t have time to take care of myself because I’m so busy with work and family. What should I do? & 4 \\
        & I have a mix of physical and mental health concerns and I don’t know how to prioritize them. & X \\
        & I am easily distracted and have trouble focusing on what to do. & 1 \\
        
        \Xhline{2\arrayrulewidth}
        Other & (Write your own) & 1 \\
        
        \Xhline{2\arrayrulewidth}
    \end{tabular}
    } 
    \caption{\textbf{Scenario List.} This table presents the list of scenarios that were presented to users to choose from.}
    \label{tab:scenario-list}
\end{table*}

\FloatBarrier
\subsection{Inter-Class Correlation to Compare Ratings}

This table contains the inter-class correlation for all questions that were rated by users, experts, and the auto-rater.
\begin{table}[h]
\centering
\begin{tabular}{rccc}
\toprule[1.5pt]
 & E. vs A. & E. vs U. & A. vs U. \\
\midrule
CUQ & 0.22 & 0.52 & 0.21 \\
WRI & 0.23 & 0.48 & 0.16 \\
CEMI & -0.09 & 0.25 & 0.26 \\
SMART & 0.34 & 0.48 & 0.41 \\ \hdashline
CUQ1 & 0.43 & 0.41 & 0.27 \\
CUQ2 & 0.09 & 0.45 & 0.06 \\
CUQ3 & 0.33 & 0.34 & 0.32 \\
CUQ4 & -0.5 & 0.22 & -0.39 \\
CUQ5 & 0.21 & 0.43 & 0.19 \\
CUQ6 & 0.43 & 0.5 & 0.39 \\
WRI1 & 0.29 & 0.39 & 0.22 \\
WRI2 & 0.36 & 0.34 & 0.23 \\
WRI3 & 0.05 & 0.38 & 0.04 \\
WRI4 & 0 & 0.45 & 0.03 \\
CEMI1 & 0.25 & 0.33 & 0.33 \\
CEMI2 & 0.1 & 0.12 & 0.3 \\
CEMI3 & -0.25 & 0.13 & 0.06 \\
CEMI4 & -0.28 & 0.18 & 0.17 \\
SMART1 & 0.29 & 0.41 & 0.39 \\
SMART2 & 0.02 & 0.22 & 0.25 \\
SMART3 & 0.19 & 0.37 & 0.19 \\
SMART4 & 0.26 & 0.4 & 0.25 \\
SMART5 & 0.45 & 0.39 & 0.3 \\
\bottomrule[1.5pt]
\end{tabular}
\caption{Intraclass Correlation Coefficients (ICC) between experts (E), users (U) and auto evals (A). Positive numbers (>0.5) signify moderate agreement.}
\label{tab:my_table}
\end{table}
\newpage

\subsection{Conversation Sample: \standout{Expert-Facilitative}}
\label{app:conv-sample}

\noindent\textbf{\color{blue} User:} 
\begin{tcolorbox}[colback=usercolor]
I feel overwhelmed by social media. How can I set healthier boundaries?
\end{tcolorbox}

\noindent\textbf{\color{darkgreen} Coach:} 
\begin{tcolorbox}[colback=coachcolor]
It's completely understandable to feel that way! A lot of people are overwhelmed by social media. To help you set some healthier boundaries, can you tell me a bit more about *why* you feel overwhelmed? Is it the amount of time you spend on it, the content you're seeing, or something else entirely? For example, do you feel pressure to keep up, or is it FOMO (fear of missing out)?
\end{tcolorbox}

\noindent\textbf{\color{blue} User:} 
\begin{tcolorbox}[colback=usercolor]
There's quite a few factors. I often notice myself automatically checking it in moments when I could be doing something else, like pursuing my creative hobbies. I'll get home after work, and after dinner and other responsibilities, will just default to spending the evening watching videos on the socials. But, that's also how I keep in touch with my friends. I'm also a DJ, so I need to use social media to keep up with my own shows, and the shows that my friends are doing.
\end{tcolorbox}

\noindent\textbf{\color{darkgreen} Coach:} 
\begin{tcolorbox}[colback=coachcolor]
So it sounds like you're struggling with balancing social media with other things you enjoy, like your hobbies, and also managing its role in your social life and work as a DJ. Do you feel like you're losing time for your hobbies because of social media, or is it more about wanting to be more intentional with your time in general? Or perhaps something else?
\end{tcolorbox}

\noindent\textbf{\color{blue} User:} 
\begin{tcolorbox}[colback=usercolor]
Yes, I think I am losing time for my hobbies due to socials. I also do want to be intentional with my time in general, but generally suck at time management. I think it also affects my mood as well - the more time I spend on social media, the less "good" I feel.
\end{tcolorbox}

\noindent\textbf{\color{darkgreen} Coach:} 
\begin{tcolorbox}[colback=coachcolor]
It's common to feel your mood shift based on how you spend your time. Since you mentioned losing time for hobbies due to social media, and wanting to be more intentional with your time, shall we start there? We can tackle the time management piece and see if that helps with the mood aspect too. What do you think?
\end{tcolorbox}

\noindent\textbf{\color{blue} User:} 
\begin{tcolorbox}[colback=usercolor]
That would be great! Just a heads up, I already have a time limit on the apps enabled (1 hour), and Instagram is hidden from my home screen, so I have to search for it when I type. I often go over this time limit and allow more time.
\end{tcolorbox}

\noindent\textbf{\color{darkgreen} Coach:} 
\begin{tcolorbox}[colback=coachcolor]
Okay, so you already have some boundaries in place, like the time limit and hiding the app, which is a great start! It sounds like sometimes you extend that time limit though. Do you typically extend it for a specific reason, like connecting with friends, or does it happen more randomly? Knowing that might help us figure out how to make the existing limit work better for you.
\end{tcolorbox}

\noindent\textbf{\color{blue} User:} 
\begin{tcolorbox}[colback=usercolor]
It often starts with wanting to look something up that I remember, or wanting to respond to someone, or just watch one person in particular. But then once you get on the app, it often makes me forget why I came to it in the first place and I get distracted by the content.
\end{tcolorbox}

\noindent\textbf{\color{darkgreen} Coach:} 
\begin{tcolorbox}[colback=coachcolor]
Start by acknowledging that you've already taken positive steps by setting time limits and hiding the app. Since you find yourself extending those limits when you go on social media to look something up or respond to someone, try being more mindful of your initial intention. Before opening the app, write down *exactly* what you need to do. Once you're on the app, refer to your note and stick to that task. After you've completed it, close the app immediately. This will help you avoid getting sidetracked by other content. Additionally, since you mentioned struggling with time management in general, consider setting aside specific times in your day dedicated to your hobbies. Treat these hobby times like important appointments and schedule them into your day. This can help you prioritize them and ensure they don't get overtaken by social media. Is there anything else you'd like to explore or any other areas you'd like to go into more depth on?
\end{tcolorbox}
\newpage

\subsection{Conversation Sample: \standout{Base-Facilitative}}

\noindent\textbf{\color{blue} User:} 
\begin{tcolorbox}[colback=usercolor]
I feel overwhelmed by social media. How can I set healthier boundaries?
\end{tcolorbox}

\noindent\textbf{\color{darkgreen} Coach:} 
\begin{tcolorbox}[colback=coachcolor]
It's completely understandable to feel overwhelmed by social media.  What specifically about social media makes you feel overwhelmed? Is it the amount of time you spend on it, the content you see, or something else?
\end{tcolorbox}

\noindent\textbf{\color{blue} User:} 
\begin{tcolorbox}[colback=usercolor]
I find myself losing time to social media. I don't have a good practice in my hobbies, because I will often get stuck or sucked in to the social media apps at night, when I could be working on my creative hobbies. As a result, I see less progress on those hobbies.
\end{tcolorbox}

\noindent\textbf{\color{darkgreen} Coach:} 
\begin{tcolorbox}[colback=coachcolor]
It's great that you've identified that social media is taking away from your hobbies.  Do you have a sense of how much time you're spending on social media each day or each week?  Is it a consistent amount of time, or does it vary?
\end{tcolorbox}

\noindent\textbf{\color{blue} User:} 
\begin{tcolorbox}[colback=usercolor]
I have a time limit on my phone for 1h on all the apps, but often find myself extending it. I am also a DJ, and need to use it to promote myself and keep tabs on others who are playing and having shows to support them. It also helps me feel connected to my friends. The time often varies.
\end{tcolorbox}

\noindent\textbf{\color{darkgreen} Coach:} 
\begin{tcolorbox}[colback=coachcolor]
It's good that you already have time limits set up – that's a great first step!  Since you sometimes extend those limits, maybe we can explore some ways to make it harder to do that.  Have you considered things like having a friend or family member change the password for your limits, or using an app that completely blocks social media after a certain time?  There are also website blockers for your computer if that's an issue too.  What do you think would be easiest for you to try?
\end{tcolorbox}

\noindent\textbf{\color{blue} User:} 
\begin{tcolorbox}[colback=usercolor]
sorry, 1h for each day.
\end{tcolorbox}

\noindent\textbf{\color{darkgreen} Coach:} 
\begin{tcolorbox}[colback=coachcolor]
You've already taken a great first step by setting time limits on your social media apps. Since you find yourself extending those limits, try making it harder to bypass them.  You could ask a friend or family member to set the limits and keep the password from you.  Alternatively, explore apps or website blockers that completely restrict access after your allotted time. Since you're a DJ and need social media for professional purposes, designate specific times during the day for work-related social media use, keeping it separate from your personal scrolling time.  This could involve using a separate account or device for work.  Also, consider scheduling specific times for your creative hobbies, just like you would an appointment, to prioritize them.  How do those ideas sound? Is there anything else you'd like to discuss, or would you like to explore any of these suggestions in more detail?
\end{tcolorbox}

\noindent\textbf{\color{blue} User:} 
\begin{tcolorbox}[colback=usercolor]
Since the apps remember the password entered, it seems like that wouldn't work very well - most of my social media time is on my phone. I also don't know about any phone apps that block apps entirely - can you recommend some?
\end{tcolorbox}

\noindent\textbf{\color{darkgreen} Coach:} 
\begin{tcolorbox}[colback=coachcolor]
You've already identified your core issue – spending too much time on social media at the expense of your hobbies.  You also have time limits set up, which is a fantastic start.  Since extending those limits is easy, let's explore stronger barriers.   
\begin{itemize}
\item First, research app blockers specifically designed for phones.  Search your app store for terms like "app blocker," "website blocker," "focus mode," or "digital wellbeing."  Read reviews and find one that suits your needs.  Many offer features like strict time limits that can't be overridden and even the ability to block specific apps during certain hours.  

\item Second, since you're a DJ and need social media professionally, create a separate account or, ideally, use a separate device solely for work-related social media. This compartmentalization will help you mentally separate work from leisure and prevent bleed-over into personal scrolling.  

\item Third, schedule dedicated time for your creative hobbies. Treat these hobby sessions like important appointments, blocking out time in your calendar and sticking to it. This will help prioritize your hobbies and make it less likely that social media will encroach on that time.  

\item Finally, consider establishing a "social media-free" zone in your home, like your bedroom or creative workspace.  This physical separation can help create a mental barrier and reduce the temptation to mindlessly scroll.  
  
\end{itemize}

Is there anything else you'd like to know, or would you like to go into more depth on any of these suggestions?
\end{tcolorbox}

\noindent\textbf{\color{blue} User:} 
\begin{tcolorbox}[colback=usercolor]
Where would you recommend my "social media-free" zone in my home be?
\end{tcolorbox}

\noindent\textbf{\color{darkgreen} Coach:} 
\begin{tcolorbox}[colback=coachcolor]
You've already identified that excessive social media use is impacting your hobbies, and you're looking to create a "social media-free" zone, which is a great idea!  I'd recommend establishing this zone in the area where you pursue your creative hobbies. This could be a dedicated room, a specific corner of a room, or even just your desk.  The key is to associate that physical space with focused, screen-free activity.  If your bedroom is where you tend to unwind and sleep, making it a social media-free zone can also improve your sleep hygiene.  Avoid bringing your phone or other devices with social media apps into this designated space.  The goal is to create a physical and mental separation between your creative pursuits and the distractions of social media.  

Is there anything else you'd like to know, or would you like to go into more depth on any of these suggestions?
\end{tcolorbox}

\newpage
\subsection{Conversation Sample: \standout{Expert-Interrogative}}
\noindent\textbf{\color{blue} User:} 
\begin{tcolorbox}[colback=usercolor]
I feel overwhelmed by social media. How can I set healthier boundaries?
\end{tcolorbox}

\noindent\textbf{\color{darkgreen} Coach:} 
\begin{tcolorbox}[colback=coachcolor]
What aspects of social media are making you feel overwhelmed?
\end{tcolorbox}

\noindent\textbf{\color{blue} User:} 
\begin{tcolorbox}[colback=usercolor]
I find myself losing time to social media. Social media can feel really addicting. I don't have a good practice in my hobbies, because I will often get stuck or sucked into the social media apps at night, when I could be working on my creative hobbies. As a result, I see less progress on those hobbies, and lots of time just watching videos. I spend too much time on it, despite the 1h limit I have set for myself on the app.
\end{tcolorbox}

\noindent\textbf{\color{darkgreen} Coach:} 
\begin{tcolorbox}[colback=coachcolor]
It sounds like you want to spend more time on your hobbies and less time on social media, even though you find it addicting. Is that right?
\end{tcolorbox}

\noindent\textbf{\color{blue} User:} 
\begin{tcolorbox}[colback=usercolor]
yep
\end{tcolorbox}

\noindent\textbf{\color{darkgreen} Coach:} 
\begin{tcolorbox}[colback=coachcolor]
Okay, cool. We can definitely work on that! What have you already tried, besides the app limits, to curb your social media use? For example, have you tried deleting apps, putting your phone in another room, or having a friend hold you accountable?
\end{tcolorbox}

\noindent\textbf{\color{blue} User:} 
\begin{tcolorbox}[colback=usercolor]
I did hide the apps from the home screen, which means I have to type them in the search to access them. This slows me down a bit, and I think has helped me not automatically open the app, but doesn't help with the time warp that sometimes happens at night, where I open it and then lose track of time.
\end{tcolorbox}

\noindent\textbf{\color{darkgreen} Coach:} 
\begin{tcolorbox}[colback=coachcolor]
So, hiding the apps helped a little, but hasn't totally solved the late-night time warp issue. Do you have a sense of *why* you're opening the apps at night? Are you bored, stressed, or just looking for a distraction? Or something else entirely?
\end{tcolorbox}

\noindent\textbf{\color{blue} User:} 
\begin{tcolorbox}[colback=usercolor]
Yeah, I think I will often open them if I have some downtime and want to relax. I think a distraction, and wanting to feel close to my friends. I'm also a DJ, and open it to post about my own shows or keep up with my friends who DJ and post.
\end{tcolorbox}

\noindent\textbf{\color{darkgreen} Coach:} 
\begin{tcolorbox}[colback=coachcolor]
You mentioned that hiding the apps helped a little, but you still find yourself getting sucked into social media at night. Since you open social media apps to relax, connect with friends, and keep up with your DJ community, try replacing that habit with alternatives. When you have downtime and want to relax, perhaps try reading a book, listening to music, or spending time on one of your creative hobbies. To connect with friends and your DJ community, consider scheduling regular phone calls or in-person meetups. Since you're already aware of the time warp effect, try setting an alarm for a specific time to put your phone away in the evening, perhaps in another room, to create a barrier and remind you to engage in other activities.
\end{tcolorbox}

\noindent\textbf{\color{blue} User:} 
\begin{tcolorbox}[colback=usercolor]
That is what I want to do, spend more time on my creative hobbies instead of on social media.
\end{tcolorbox}

\noindent\textbf{\color{darkgreen} Coach:} 
\begin{tcolorbox}[colback=coachcolor]
To reclaim your time for creative hobbies and curb late-night social media use, try replacing your social media habit with alternative activities. When you want to relax, pick up a book, listen to music, or work on your creative hobbies instead of opening social media apps. To connect with friends and your DJ community, schedule regular phone calls or in-person meetups. Finally, set an alarm to remind yourself to put your phone away in a separate room each evening, creating a physical barrier to mindless scrolling and prompting you to engage with your hobbies.
\end{tcolorbox}

\noindent\textbf{\color{blue} User:} 
\begin{tcolorbox}[colback=usercolor]
That's a good suggestion, thanks.
\end{tcolorbox}

\noindent\textbf{\color{darkgreen} Coach:} 
\begin{tcolorbox}[colback=coachcolor]
To create a barrier between you and your phone at night, put your phone in a separate room, ideally outside of your bedroom. A good spot might be the living room or kitchen. The key is to make it slightly inconvenient to access so you're less tempted to mindlessly pick it up.
\end{tcolorbox}

\newpage

\subsection{Conversation Sample: \standout{Base-Interrogative}}
\noindent\textbf{\color{blue} User:} 
\begin{tcolorbox}[colback=usercolor]
I feel overwhelmed by social media. How can I set healthier boundaries?
\end{tcolorbox}

\noindent\textbf{\color{darkgreen} Coach:} 
\begin{tcolorbox}[colback=coachcolor]
What aspects of social media are contributing the most to this feeling of being overwhelmed?
\end{tcolorbox}

\noindent\textbf{\color{blue} User:} 
\begin{tcolorbox}[colback=usercolor]
It's pretty addicting, and takes time away from my creative hobbies. I often notice a time-warp sometimes, when I sit down for the night, and then lose an hour or two just to watching videos - even though I already have a 1h time limit on it.
\end{tcolorbox}

\noindent\textbf{\color{darkgreen} Coach:} 
\begin{tcolorbox}[colback=coachcolor]
It's completely understandable how that can happen. It sounds like you've already taken a good first step by setting time limits. Do you find yourself hitting that 1-hour limit often? What happens when you do? Does the app lock you out, or is it just a notification?
\end{tcolorbox}

\noindent\textbf{\color{blue} User:} 
\begin{tcolorbox}[colback=usercolor]
Social media also makes me have unrealistic expectations for how fun or interesting my life should be, I think. Yes, I hit the 1h limit most days. The time limit stops you from opening the app, but has a "snooze" feature to allow another 1 minute or 15 minutes, or can remove the block for the day. I also have the app hidden from my home screen, which prevents me from automatically opening it, and helps to slow me down, since I have to type it in to find it.
\end{tcolorbox}

\noindent\textbf{\color{darkgreen} Coach:} 
\begin{tcolorbox}[colback=coachcolor]
What happens after you type the app name in? Do you still open it even though you're actively trying to avoid it?
\end{tcolorbox}

\noindent\textbf{\color{blue} User:} 
\begin{tcolorbox}[colback=usercolor]
Yes, I will often still open it if I'm over the 1h limit. I will often want to open the app to keep up with my friends, post something about myself (I am a DJ, and so use it for promotion), or look something up that I had seen before. It will often distract me from my actual goal, and keep me involved in the app watching videos, etc.
\end{tcolorbox}

\noindent\textbf{\color{darkgreen} Coach:} 
\begin{tcolorbox}[colback=coachcolor]
What specifically keeps you involved in the app, even when you know you should be doing something else?
\end{tcolorbox}

\noindent\textbf{\color{blue} User:} 
\begin{tcolorbox}[colback=usercolor]
My energy levels are often low at the end of the day after work. I will sometimes open the app to relax, but then will get sucked in watching funny or interesting content. It's easy to do, and sometimes the creative hobbies are "hard" to do, and require setup, learning, etc., even though they are much more rewarding in the long run.
\end{tcolorbox}

\noindent\textbf{\color{darkgreen} Coach:} 
\begin{tcolorbox}[colback=coachcolor]
It’s great that you recognize the long-term rewards of your creative hobbies! It sounds like you’re using social media as a way to unwind after work, which is understandable given the lower energy levels. Have you considered trying any other relaxing activities before turning to social media? For example, a quick walk, listening to music, or even just a few minutes of meditation could help recharge you before diving into your hobbies.
\end{tcolorbox}

\noindent\textbf{\color{blue} User:} 
\begin{tcolorbox}[colback=usercolor]
I have, but social media is often a default action for me. It also can help me feel connected to my friends and what's going on in my city. I haven't considered trying to intentionally relax, though.
\end{tcolorbox}

\noindent\textbf{\color{darkgreen} Coach:} 
\begin{tcolorbox}[colback=coachcolor]
You've already taken some great steps by setting time limits and hiding the app. Since your energy is low after work and you tend to default to social media, try incorporating a short, intentional relaxation activity *before* you even consider opening the app. This could be a quick walk, listening to music, a few minutes of meditation, or anything else that helps you unwind. This will hopefully give you the energy boost you need to pursue your more rewarding creative hobbies instead of getting drawn into social media. Also, since you value the connection aspect of social media, perhaps schedule specific times for it, so you can still connect with friends and events without it derailing your entire evening.
\end{tcolorbox}

\noindent\textbf{\color{blue} User:} 
\begin{tcolorbox}[colback=usercolor]
do you think I should schedule it towards the beginning or end of my night? And when should I try to do my intentional relaxing activity - right when I get home? What should I check out to get some ideas for the activity
\end{tcolorbox}

\noindent\textbf{\color{darkgreen} Coach:} 
\begin{tcolorbox}[colback=coachcolor]
You should try incorporating a short, intentional relaxation activity right when you get home from work, *before* you even consider opening social media. To get ideas for relaxing activities, explore options like short walks, listening to music, a few minutes of meditation, or anything else that helps you unwind and recharge. You can also search online for "relaxation techniques" or "stress reduction activities" for more inspiration. By intentionally relaxing first, you'll hopefully boost your energy levels enough to pursue your creative hobbies instead of defaulting to social media. Additionally, schedule specific times for social media to maintain connection with friends and events without it taking over your evenings.
\end{tcolorbox}

\newpage
\subsection{Conversation Sample: \standout{Directive}}
\noindent\textbf{\color{blue} User:} 
\begin{tcolorbox}[colback=usercolor]
I feel overwhelmed by social media. How can I set healthier boundaries?
\end{tcolorbox}

\noindent\textbf{\color{darkgreen} Coach:} 
\begin{tcolorbox}[colback=coachcolor]
Start small. Choose one platform to limit initially. Set daily time limits, disable notifications, or schedule specific times for checking it. Be mindful of *why* you're reaching for your phone – are you bored, procrastinating, or genuinely connecting? Find alternative activities you enjoy.
\end{tcolorbox}

\noindent\textbf{\color{blue} User:} 
\begin{tcolorbox}[colback=usercolor]
I do actually already have a limit to my social media time - it's at 1h, but I often find myself going over. The mindfulness is the issue here - I often will reach for my phone or social media as a relaxing break, but it can feel addicting and "time warp" me through to the end of my night.
\end{tcolorbox}

\noindent\textbf{\color{darkgreen} Coach:} 
\begin{tcolorbox}[colback=coachcolor]
It's common to fall into that "time warp" trap! Since relaxation is your goal, let's find replacements. What activities help you unwind *without* your phone? Think hobbies, spending time with loved ones, being in nature, etc. We can then build those into your routine so you have healthy alternatives readily available.
\end{tcolorbox}

\noindent\textbf{\color{blue} User:} 
\begin{tcolorbox}[colback=usercolor]
My goal would be to spend more time on my creative hobbies, but it's often easier to spend time on social media than taking the time to set up, and then clean up, a creative hobby. These are more rewarding than being on social media, however.
\end{tcolorbox}

\noindent\textbf{\color{darkgreen} Coach:} 
\begin{tcolorbox}[colback=coachcolor]
It's understandable. Prep and cleanup *can* be deterrents. Try minimizing those barriers. Dedicate a specific space for your hobby so setup is minimal. Have easy cleanup solutions ready, like wipes or a designated "messy" area. Schedule short creative bursts, even 15 minutes, to build momentum without feeling overwhelmed.
\end{tcolorbox}

\noindent\textbf{\color{blue} User:} 
\begin{tcolorbox}[colback=usercolor]
Good suggestion, thanks.
\end{tcolorbox}

\noindent\textbf{\color{darkgreen} Coach:} 
\begin{tcolorbox}[colback=coachcolor]
How about we pick one hobby and brainstorm ways to make it more accessible this week? What comes to mind?
\end{tcolorbox}

\noindent\textbf{\color{blue} User:} 
\begin{tcolorbox}[colback=usercolor]
I need to make a zine for some friends that has quotes from a book we've been reading in book club, called "The Artist's Way." It's our final project so to speak, as it will be the last week of the book club. It's due on Monday, and it's already Friday. There should be 8 total pages. Each page will have one quote from the book that I like, and a drawing by me.
\end{tcolorbox}

\noindent\textbf{\color{darkgreen} Coach:} 
\begin{tcolorbox}[colback=coachcolor]
Okay, the zine is a perfect starting point! Let's break it down. Eight pages isn't overwhelming. Could you choose your eight quotes today? Then aim for two pages per day – one quote and drawing each. Have all your supplies in one place. No cleanup until Sunday, deal?
\end{tcolorbox}

\noindent\textbf{\color{blue} User:} 
\begin{tcolorbox}[colback=usercolor]
Deal!
\end{tcolorbox}

\noindent\textbf{\color{darkgreen} Coach:} 
\begin{tcolorbox}[colback=coachcolor]
Great! Let's check in tomorrow to see how those first two pages are coming along. What time works best for a quick message?
\end{tcolorbox}

\noindent\textbf{\color{blue} User:} 
\begin{tcolorbox}[colback=usercolor]
Noon.
\end{tcolorbox}

\noindent\textbf{\color{darkgreen} Coach:} 
\begin{tcolorbox}[colback=coachcolor]
Perfect. Noon tomorrow it is. I'm confident you'll make great progress! See you then.
\end{tcolorbox}

\end{document}